\pdfoutput=1
\documentclass{article}

\usepackage{PRIMEarxiv}

\usepackage[utf8]{inputenc} 
\usepackage[T1]{fontenc}    
\usepackage[table]{xcolor}
\usepackage{hyperref}       
\usepackage{url}            
\usepackage{booktabs}       
\usepackage{amsfonts}       
\usepackage{nicefrac}       
\usepackage{microtype}      
\usepackage{lipsum}
\usepackage{graphicx}
\usepackage{tabularx}
\usepackage{tikz}
\usepackage{placeins}
\usepackage{float}
\usepackage{booktabs}
\usepackage{tabularx}
\usepackage{array}
\usepackage{ragged2e}
\usepackage{amsmath}

\usepackage{tikz}

\newcolumntype{Y}{>{\RaggedRight\arraybackslash}X}

\usepackage{booktabs}
\usepackage{tabularx}
\usepackage{array}
\usepackage{makecell}
\usetikzlibrary{arrows.meta,backgrounds,calc,fit,positioning,shapes.geometric}
\graphicspath{{media/}}     

\providecommand{\aifourr}{AI4R}
\providecommand{\tool}[1]{\texttt{#1}}
\definecolor{BaizeInk}{HTML}{24313A}
\definecolor{BaizeTeal}{HTML}{167D89}
\definecolor{BaizeBlue}{HTML}{356A91}
\definecolor{BaizeCoral}{HTML}{C65E57}
\definecolor{BaizeGold}{HTML}{B68A2C}
\definecolor{BaizeMist}{HTML}{F3F6F7}
\definecolor{BaizeLine}{HTML}{C9D2D6}

\title{Towards a new paradigm of scientific discovery with socialized artificial intelligence
} 

\newcommand{\inst}[1]{\textsuperscript{#1}}

\author{\\
Xinjie Yao\inst{1, 2, 3, *},
Xingxin Xu\inst{5, *},
Xiyuan Gao\inst{5, *},
Zhoupeng Guo\inst{4, *},
Kunlong Yang\inst{6},
Dengyu Zhao\inst{5},
Siqi Zhao\inst{5},\\
Zhihe Fan\inst{5},
Yichen Dong\inst{5},
Xin Li\inst{4},
Jiekang Feng\inst{5},
Jiahe Wu\inst{5},
Sen Wang\inst{4},
Beiming Yu\inst{5},\\
Kejia Zhao\inst{4},
Ruipu Zhao\inst{5},
Jiaqi Zhou\inst{5},
Heyang Li\inst{5},
Jianjun Chen\inst{5},
Anbo Dai\inst{7},\\
Xin Liu\inst{1, 7, 8, $\dagger$},
Zhengtao Yu\inst{2, 3, $\dagger$},
Qinghua Hu\inst{5, $\dagger$},
Pengfei Zhu\inst{4, 5, $\dagger$}
\\
\\
\inst{1} Baize Research\\
\inst{2} Faculty of Information Engineering and Automation, Kunming University of Science and Technology\\
\inst{3} Yunnan Key Laboratory of Artificial Intelligence, Kunming University of Science and Technology\\
\inst{4} School of Automation, Southeast University\\
\inst{5} School of Artificial Intelligence, Tianjin University\\
\inst{6} School of Computer Science and Technology, Beijing Institute of Technology\\
\inst{7} SeetaCloud Technology\\
\inst{8} GPUhub Pte. Ltd.\\
\\
* Equal Contribution\\
$\dagger$ Corresponding Authors\\
\\
\href{https://www.oplclaw.com}{\texttt{https://www.oplclaw.com/}}
}

\begin{document}

\begin{tikzpicture}[remember picture,overlay]
\node[anchor=north, yshift=-1cm] 
at (current page.north)
{
\raisebox{-0.5\height}{%
    \includegraphics[height=1.1cm]{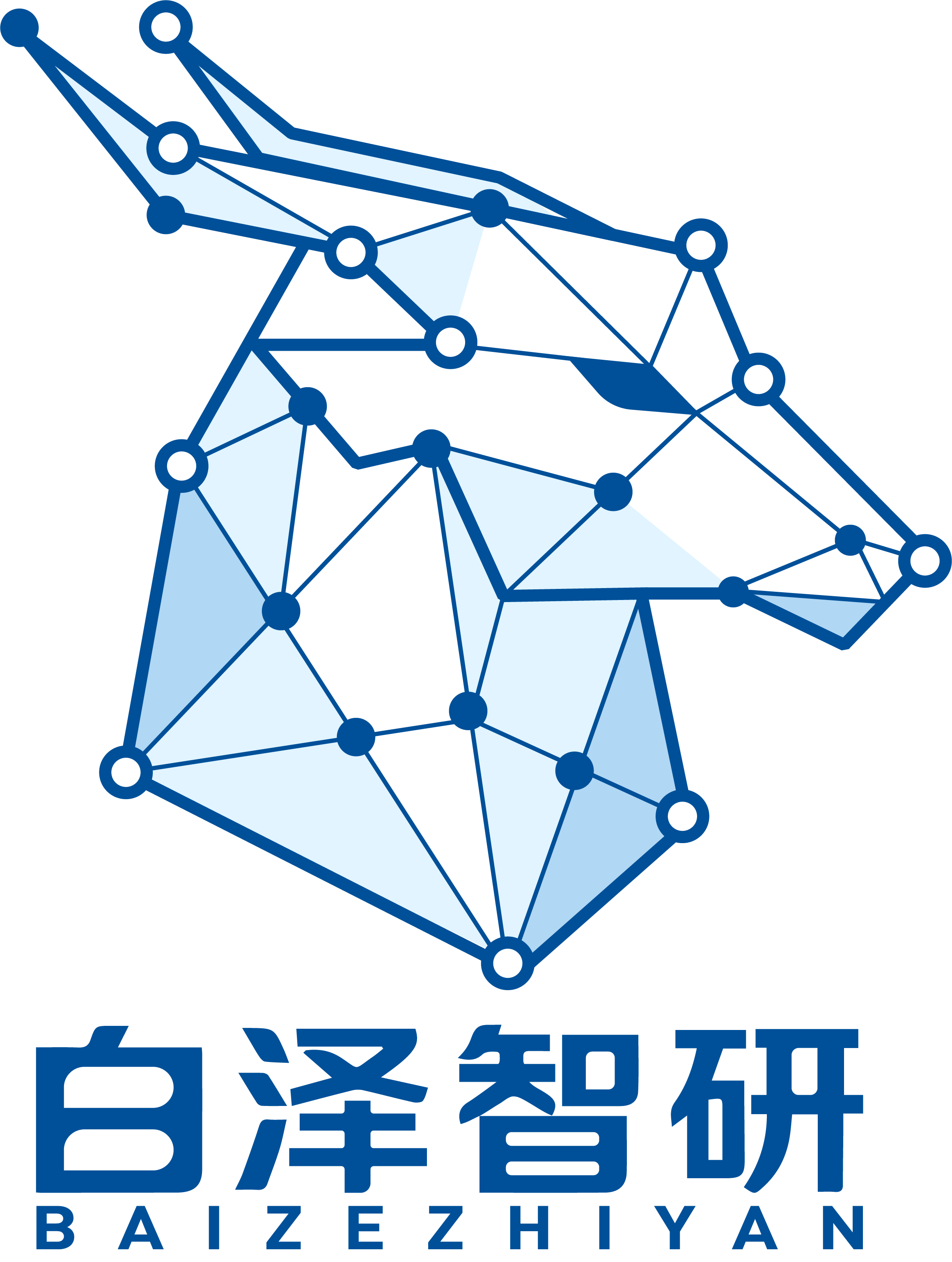}
}
\quad
\raisebox{-0.5\height}{%
    \includegraphics[height=1.0cm]{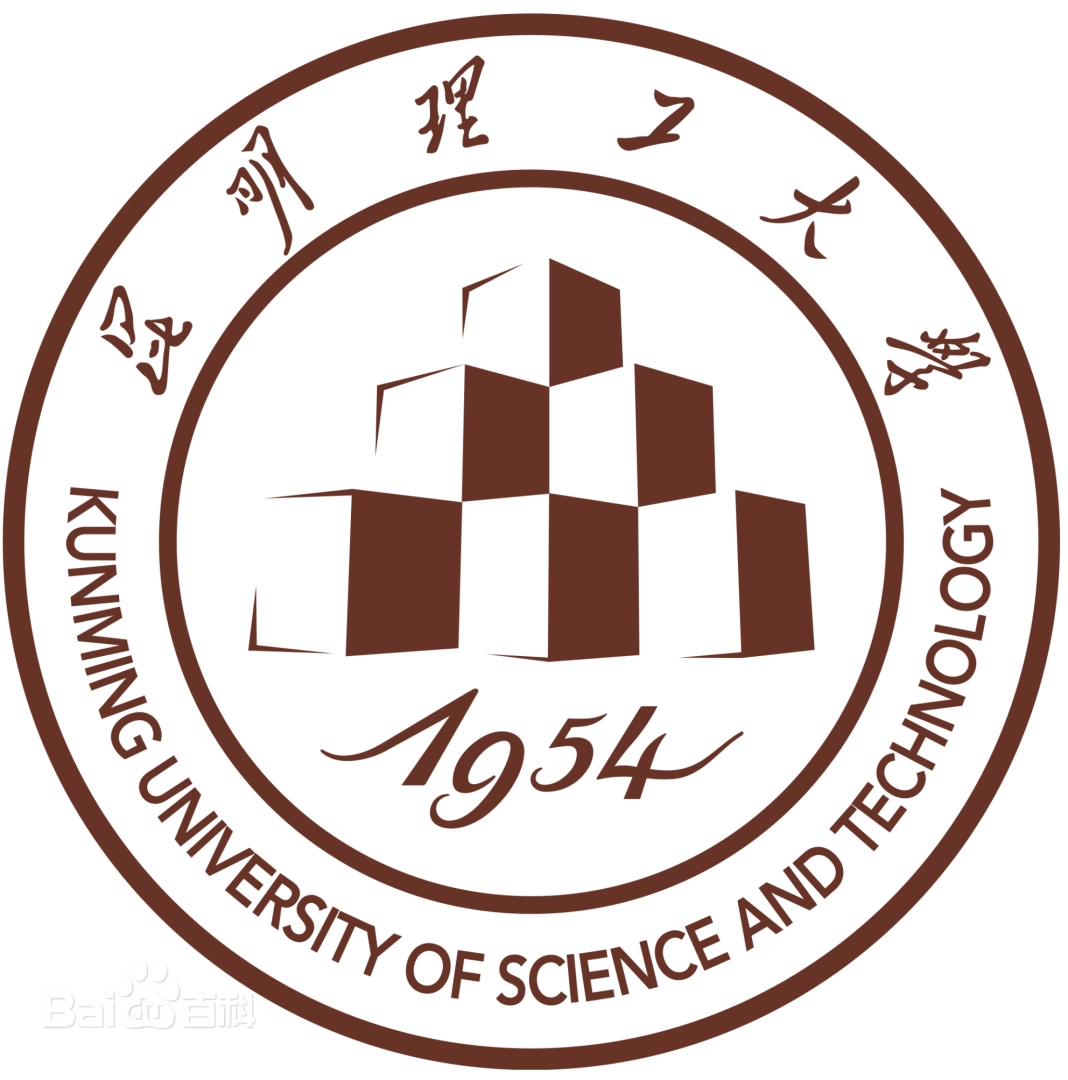}
}
\quad
\raisebox{-0.5\height}{%
    \includegraphics[height=1.0cm]{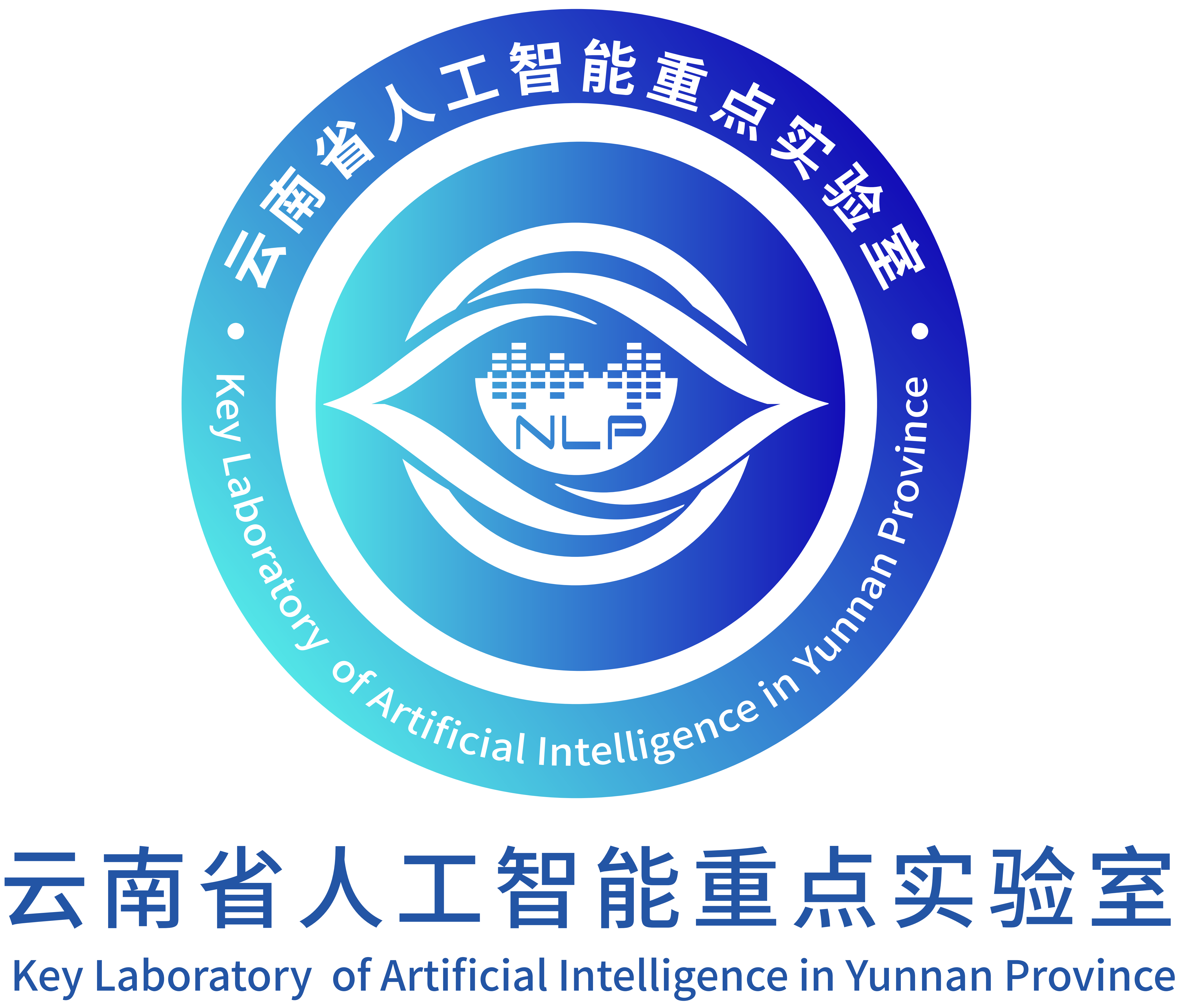}
}
\quad
\raisebox{-0.5\height}{%
    \includegraphics[height=1.0cm]{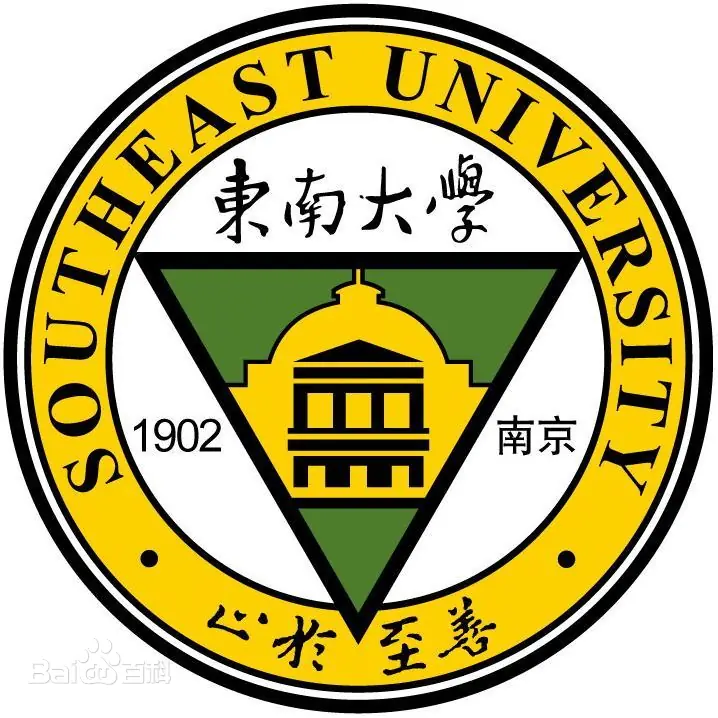}
}
\quad
\raisebox{-0.5\height}{%
    \includegraphics[height=1.0cm]{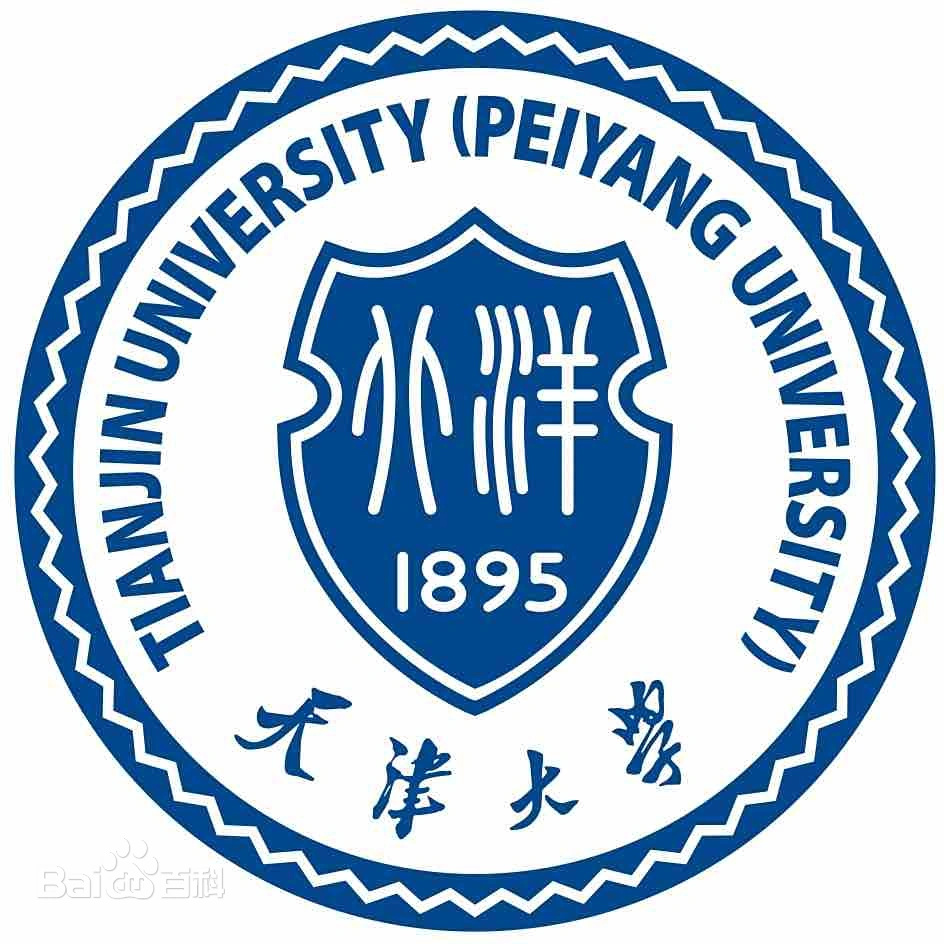}
}
\quad
\raisebox{-0.5\height}{%
    \includegraphics[height=1.0cm]{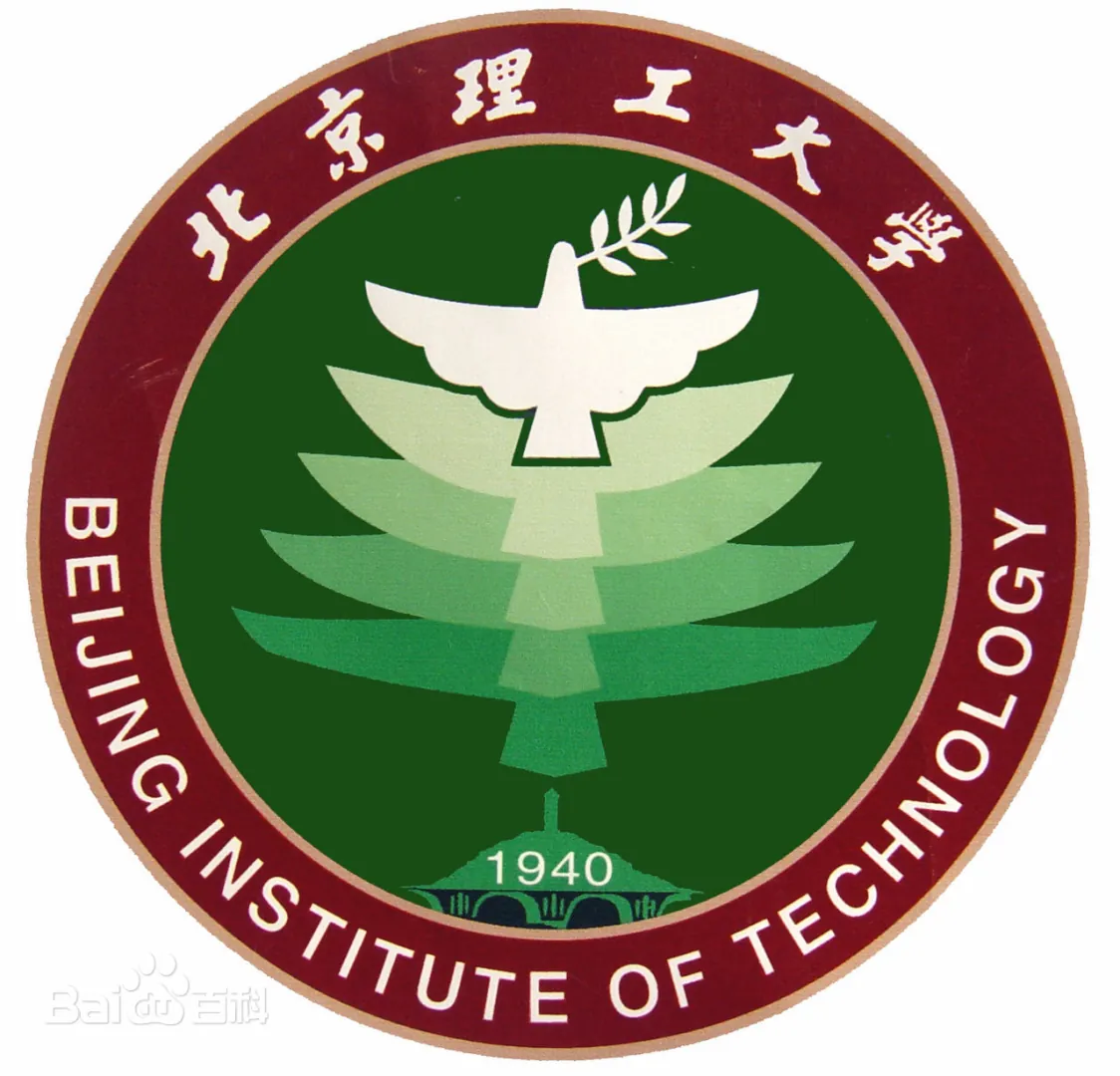}
}
\quad
\raisebox{-0.5\height}{%
    \includegraphics[height=0.65cm]{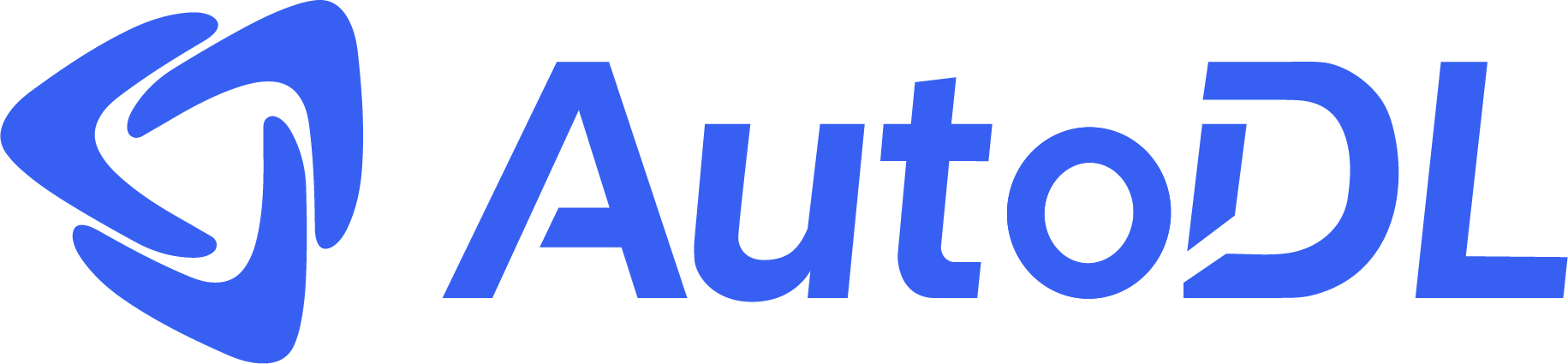}
}
\quad
\raisebox{-0.5\height}{%
    \includegraphics[height=0.85cm]{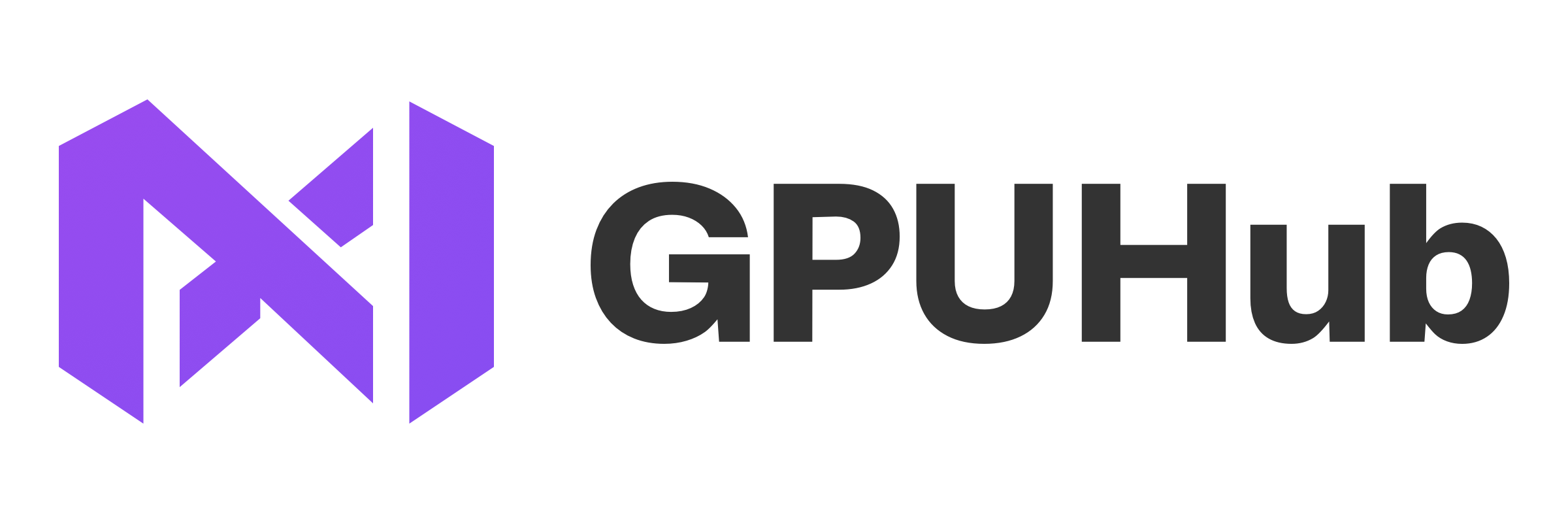}
}
};
\end{tikzpicture}

\maketitle

\begin{abstract}

Scientific discovery has advanced through successive transformations in the organization of knowledge. Observation and experimentation established the empirical foundations of science. Theory made it possible to derive general principles from particular phenomena. Computation extended inquiry into systems beyond direct observation, while data-intensive methods opened new spaces of pattern and prediction. Science now confronts a different frontier. The central challenge is no longer simply to produce more information, but to organize expanding knowledge, reasoning, and evidence into a coherent process of discovery.
Here, we introduce Bridging Literature, Agents, and Zero-gap Experimentation (BLAZE), a paradigm of socialized scientific intelligence. BLAZE conceives AI not as an assistant for isolated research tasks, but as an organizational infrastructure for scientific discovery. It connects persistent knowledge, collective reasoning, empirical validation, and human judgment within a continuous research lifecycle, transforming fragmented activities into a cumulative process of inquiry, criticism, and revision.
The central premise of BLAZE is that scientific intelligence does not arise from computation alone. It emerges from the sustained interaction among knowledge, hypotheses, experiments, and collective verification. By organizing humans and machines within a shared scientific process, BLAZE makes discovery more traceable, reproducible, and cumulative while preserving human creativity, judgment, and responsibility.
Socialized scientific intelligence may provide a foundation for the next era of science. Its purpose is not to replace human discovery, but to extend the scale, depth, and continuity of collective scientific inquiry.

\end{abstract}


\section{Introduction}

\subsection{Scientific Discovery in the Age of AI}

The history of science is shaped not simply by the emergence of increasingly powerful instruments, but by recurring transformations in how knowledge is recorded, organized, validated and shared~\cite{kuhn1970structure}. Systematic observation and durable empirical records established enduring foundations for scientific inquiry, while theoretical modeling introduced complementary means of abstracting diverse phenomena into mathematical principles and extending explanation beyond what could be directly observed~\cite{hacking1983representing}. Experimental practices further strengthened this evolving knowledge system through controlled intervention, measurement and reproducibility, enabling scientific claims to be independently tested and collectively scrutinized~\cite{shapin2011leviathan}. Observation, modeling, and experimentation continue to coexist, interact and reinforce one another as complementary mechanisms of scientific reasoning~\cite{galison1997image}. From this perspective, scientific revolutions arise not only from new instruments, but also from fundamental changes in how knowledge is documented, validated and collaboratively produced.

As scientific problems grew in scale and complexity, computation expanded the range of systems and hypotheses that could be explored. Numerical simulation made it possible to study phenomena that were difficult to access through direct experimentation~\cite{anderson1972more}, while large-scale data and machine learning enabled the discovery of patterns that could not be readily expressed through existing theories~\cite{hey2009fourth}. Computation thus evolved from evaluating human-designed hypotheses to supporting representation learning, pattern discovery and search over increasingly large scientific spaces~\cite{kumar2025automating}. Yet a larger search space does not by itself determine which questions are worth pursuing, how evidence should be organized and validated, or how knowledge should remain coherent across different stages of research.

\begin{figure*}[h]
    \centering
    \includegraphics[width=0.95\textwidth]{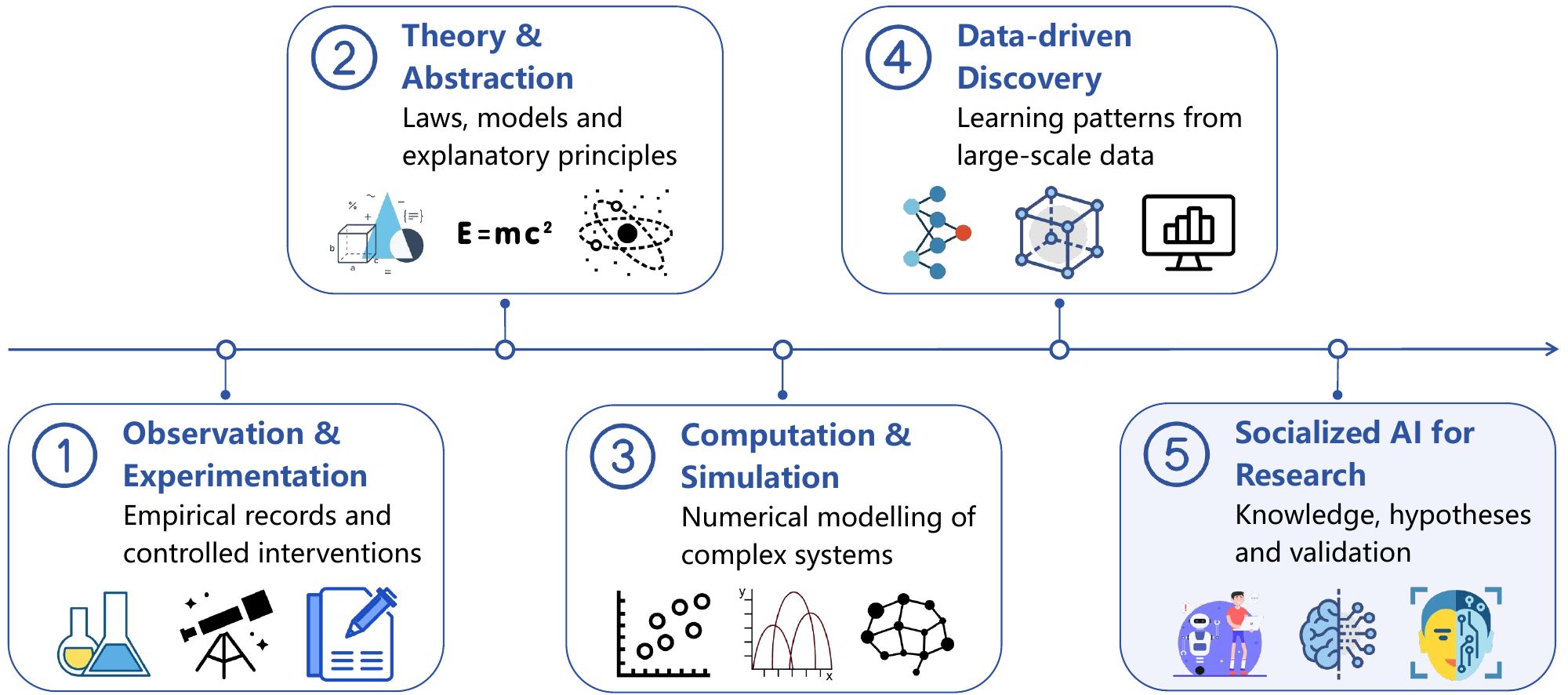}
    \caption{The evolving organization of scientific discovery.
    Scientific inquiry has progressively expanded from observation and experimentation to theoretical abstraction, computational simulation and data-driven discovery. Building on rather than replacing these established modes, AI4R is introducing an emerging organizational layer through which knowledge, hypotheses and validation can be coordinated across the scientific lifecycle.}
    \label{fig:paradigm_evolution}
\end{figure*}

As illustrated in Fig.~\ref{fig:paradigm_evolution}, scientific discovery has progressively expanded from observation, experimentation and theoretical abstraction to computational simulation and data-driven inquiry. Recent advances in foundation models and agentic systems are giving rise to an emerging organizational layer of AI for Research (AI4R). This layer concerns how knowledge, hypotheses, experiments, evidence, criticism and authority are connected and sustained across time. AI systems are increasingly capable of participating in literature synthesis, hypothesis generation, experimental planning, code execution, result interpretation and scholarly communication~\cite{hao2025research,lu2025agentic}, while progress towards end-to-end research automation demonstrates that these activities can be integrated within computational workflows~\cite{liao2024architecture,li2025build}. The central question, however, is no longer whether AI can execute individual research steps, but whether human and machine contributions can be organized into a continuous, traceable and accountable scientific process. We refer to the capacity emerging from such organization as \emph{socialized scientific intelligence}: the collective capacity that arises when human judgment, machine reasoning, experimental evidence and institutional memory are coordinated within a persistent and governed scientific process. Realizing this capacity requires more than connecting isolated tools; it depends on persistent scientific memory, structured coordination, evidence-grounded validation and explicit human responsibility~\cite{li2024ai4r,liu2026foundation}.

\subsection{Bottlenecks and the Rise of Automated Research}

Modern science is increasingly constrained not only by the availability of tools, data or expertise, but also by the difficulty of organizing rapidly expanding scientific activity. The accelerating growth of publications places an increasing cognitive burden on individual researchers, making it difficult to maintain a comprehensive and up-to-date understanding of relevant knowledge~\cite{bornmann2015growth}. At the same time, automated methods are increasing the rate at which hypotheses, analyses and scientific claims can be produced, often faster than they can be independently examined and validated. Because literature review, ideation, experimentation, writing and revision remain distributed across disconnected tools and records, these growing outputs are difficult to preserve as coherent, traceable and verifiable scientific knowledge.

Automated research systems have emerged in response to these pressures. Recent approaches have begun to integrate literature analysis, idea generation, code execution, experimental evaluation, manuscript preparation and peer review within unified pipelines~\cite{yang2026aris,li2026autosota,tang2026fars}. These systems demonstrate that substantial portions of the research workflow can be automated. However, connecting research tasks into a pipeline does not by itself resolve the organizational challenges of scientific discovery. Existing systems continue to exhibit four fundamental limitations: (1) \emph{temporary scientific memory}, in which knowledge and intermediate results are not persistently retained across research cycles; (2) \emph{broken hypothesis-evidence-claim continuity}, whereby hypotheses, experiments, evidence and conclusions remain weakly linked; (3) \emph{undifferentiated or weakly accountable agent interaction}, with limited role specialization, sustained criticism and responsibility attribution; and (4) \emph{unclear human authority boundaries}, particularly for consequential decisions concerning validation, revision and publication.

The next stage of AI4R depends not only on broader task coverage or greater autonomy, but on organizing research as a connected and inspectable process~\cite{hu2025survey}. Scientific knowledge, hypotheses, experiments, evidence and evaluation must remain linked and reusable across research cycles~\cite{kitano2021nobel,wang2023scientificNature}. We formulate this challenge through four research questions:

\begin{itemize}
    \item \textbf{Q1: Continuity.} How can an AI4R system preserve evolving scientific knowledge, including prior hypotheses, experimental histories, failed attempts and decision rationales, across successive research cycles?

    \item \textbf{Q2: Traceability.} How can hypotheses, research plans, experiments, evidence and claims be linked into a reproducible and inspectable scientific process rather than retained as disconnected outputs?

    \item \textbf{Q3: Productive disagreement.} How can role-differentiated agents engage in structured criticism, evidence-based negotiation and revision, rather than merely exchanging information or reinforcing shared biases?

    \item \textbf{Q4: Accountability.} How can AI systems provide scalable research assistance while preserving explicit human authority over scientific objectives, ethical considerations, experimental risks and final conclusions?
\end{itemize}

\subsection{BLAZE: A Socialized Infrastructure for Scientific Discovery}

\begin{figure*}[h]
    \centering
    \includegraphics[width=1\textwidth]{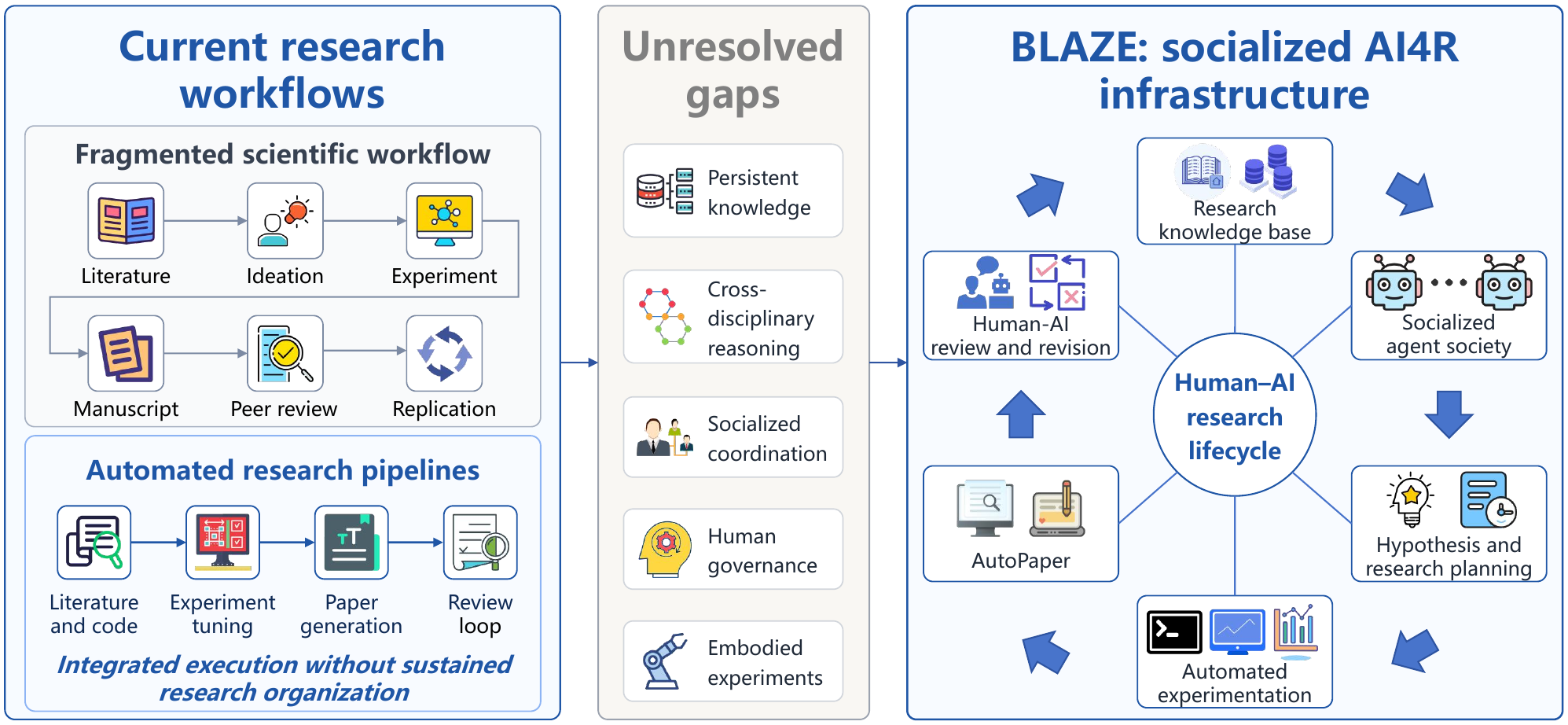}
    \caption{From fragmented research workflows to a socialized AI4R infrastructure.
    Existing workflows remain either fragmented or linearly automated, limiting the continuity, accountability and cumulative development of scientific research. BLAZE connects persistent knowledge, socialized reasoning, evidence-grounded experimentation, manuscript construction and human-AI review within a recurrent research lifecycle.}
    \label{fig:research_workflow}
\end{figure*}

We present Bridging Literature, Agents, and Zero-gap Experimentation \textbf{(BLAZE)}, a socialized AI4R infrastructure designed to organize scientific discovery as a continuous, evidence-grounded and human-governed process, as illustrated in Fig.~\ref{fig:research_workflow}. BLAZE begins with a human-specified research question, together with its scientific scope, prior literature, project materials, resource constraints and governance requirements. These inputs instantiate a persistent research object that connects hypotheses, research plans, experiment graphs, evidence, claims, reviews and human decisions throughout the project. Rather than processing literature analysis, hypothesis generation, experimentation, manuscript construction and review as independent tasks, BLAZE advances this object through guarded transitions across knowledge construction, deliberation, execution and validation. Each transition must preserve provenance, update the evidentiary state and satisfy the corresponding quality or human-approval gate; unresolved objections and insufficient evidence return the project to an earlier stage for revision or additional experimentation. The resulting outputs are therefore not limited to a manuscript, but include auditable literature records, hypothesis and decision histories, executable plans, code and configurations, experimental logs and results, claim-evidence links, review and revision records, and approval traces that remain reusable across successive research cycles.

Architecturally, BLAZE is built upon four interconnected layers that reflect the fundamental dynamics of scientific discovery. The knowledge and memory layer enables cumulative intelligence by preserving scientific context across time. The agent society layer recreates the diversity of scientific roles through structured interaction, allowing ideas to emerge, compete and evolve. The tool and execution layer closes the loop between hypotheses and evidence through reproducible experimentation. The evaluation and governance layer ensures that scientific conclusions remain grounded in evidence and guided by human responsibility. Together, these layers establish a new paradigm in which AI functions not merely as a research assistant, but as a persistent scientific organization.

Central to this architecture is a traceable \emph{Research Object} model that captures the evolution of scientific inquiry from hypotheses and research plans to experiments, evidence, claims and manuscript revisions, together with the dependencies that connect them. Rather than merely recording final outcomes, it preserves the intellectual trajectory of discovery, including hypotheses, objections, alternative explanations, failed attempts, supporting or contradictory evidence, and human decisions. Experimental processes produce not only results, but also configurations, logs and intermediate artifacts that remain explicitly linked to the hypotheses they examine and the claims they inform. Concerns raised by human and AI reviewers are transformed into structured revision or additional-experiment tasks~\cite{boiko2023autonomous,hatakeyama2025perspective}, allowing unresolved questions and evidentiary gaps to persist as part of the research state. In this view, a manuscript is not the endpoint of scientific discovery, but an interpretable projection of an evolving evidence landscape.

The empirical claims of this paper are intentionally limited. We report three completed analyses: an open-ended AutoPaper case spanning baseline reproduction, candidate screening, positive and negative results, internal review and manuscript production; a fixed-leaderboard H-EDML case examining controlled reproduction and transparently reported optimization; and an artifact-based comparison of frozen local delivery packages from BLAZE/AutoPaper, ARIS~\cite{yang2026aris}, EvoScientist~\cite{lyu2026evoscientist} and FARS~\cite{tang2026fars}, including a same-task comparison between BLAZE and ARIS on ice-shelf PINN inversion. These analyses examine lifecycle coverage, artifact traceability, protocol discipline, negative-result retention and delivery auditability. They do not establish general improvements in scientific reliability, efficiency or quality. To support broader evaluation, we further define a protocol for evidence-constrained manuscript generation and preregister a randomized controlled study involving 50 senior undergraduate, master’s and doctoral students. The study will examine whether reliability-calibrated AI feedback improves the detection of major issues without increasing false concerns or causing anchoring-induced errors in human judgment. Our contributions are summarized as follows:

\begin{itemize}
    \item We conceptualize \textbf{socialized AI4R} as a paradigm for scientific discovery, integrating persistent knowledge, agent coordination, evidence-grounded execution and human governance through a unified architecture.

    \item We propose a traceable \textbf{Research Object} that captures the evolution of scientific inquiry by linking hypotheses, plans, experiment graphs, evidence, claims, reviews and revisions into an inspectable and reusable record.

    \item We establish a \textbf{human-AI governance framework} with explicit authority boundaries, approval gates and adjudication rules, preserving human responsibility for consequential scientific decisions.

    \item We conduct an \textbf{artifact-based evaluation} on completed research cases and delivery packages, examining lifecycle coverage, provenance traceability, protocol discipline, negative-result retention and auditability.
\end{itemize}


\section{Related Work}
\subsection{AI for Science and Scientific Foundation Models}
AI for Science reduces the cost of exploring large scientific spaces by prioritizing candidates, approximating complex processes, and guiding experiments. Existing methods can be classified by where intelligence enters the discovery process: predictive foundation models learn reusable representations of scientific objects, generative search methods propose and evaluate new solutions, and closed loop discovery systems connect computational decisions with experimental feedback. These routes improve prediction, design, and experimental efficiency, but generally assume that the task, constraints, and evaluation criteria have already been specified.

(1) Predictive foundation models learn structure aware representations for proteins and materials. AlphaFold and AlphaFold 3 predict protein structures and biomolecular complexes, while GNoME expands the search for stable inorganic structures~\cite{jumper2021highly,abramson2024accurate,merchant2023scaling}. Their transferability remains limited by data coverage and domain assumptions. (2) Generative search methods use explicit evaluators to retain useful programs or algorithms, as shown by FunSearch and AlphaTensor~\cite{romera2024mathematical,fawzi2022discovering}. Their success depends on bounded search spaces and reliable objectives. (3) Closed loop systems connect predicted candidates to robotic synthesis and use observations to revise later experiments~\cite{szymanski2023autonomous}, but remain specialized to engineered domains.

Existing systems can therefore solve consequential tasks within bounded domains, but they do not decide which question matters, whether a result closes a knowledge gap, or what evidence supports a broader claim. Scientific data are often costly and heterogeneous, while the published record is biased toward positive outcomes~\cite{fanelli2012negative}. Prediction alone cannot establish explanation, realizability, or accountability. BLAZE addresses this gap by coordinating specialized models across problem selection, hypothesis formation, experimentation, evidence evaluation, writing, and criticism through shared knowledge, explicit roles, provenance tracking, and continuous verification.

\subsection{Literature Intelligence, Scientific Knowledge Bases and Automated Surveys}
Literature intelligence converts the fragmented scientific record into evidence for research decisions. Its purpose is to recover how claims, methods, experiments, and disagreements define a field, rather than merely locating papers. Existing methods can be classified into retrieval and representation methods that identify relevant documents, extraction and knowledge representation methods that structure scientific content, and synthesis and planning methods that combine sources to produce surveys or research directions. This progression moves from finding documents to organizing evidence and using it throughout research.

(1) Retrieval and representation methods use text, citation graphs, and multifaceted queries. SPECTER learns citation informed document embeddings, while DORIS-MAE evaluates retrieval under complex scientific constraints~\cite{cohan2020specter,wang2023scientific}; topical similarity still does not ensure evidential relevance. (2) Extraction and knowledge representation methods recover structured records or connect claims with evidence and challenges~\cite{dagdelen2024structured,clark2014micropublications}, but remain sensitive to terminology and source complexity. (3) Synthesis and planning methods distinguish core citations, perform citation attribution, generate surveys, and iteratively propose ideas~\cite{hao2024hlm,press2024citeme,wang2024autosurvey,baek2025researchagent}; fluent outputs can still conceal weak support, conflicting protocols, and negative results.

The field has progressed from topical search to structured synthesis, but precise claim support, persistent project memory, and cross stage coordination remain limited. Systems rarely retain rejected assumptions, failed experiments, reviewer interventions, or decision rationales, and retrieval, ideation, experimentation, writing, and review are usually separate interactions. BLAZE addresses these gaps with a provenance aware knowledge layer that connects literature, private documents, code, experimental logs, claims, reviews, and revisions. Different agents can inspect the same evidence for novelty, experiment design, manuscript support, and criticism, making literature intelligence persistent research infrastructure.

\subsection{Scientific Hypothesis Generation and Idea Discovery Agents in AI4R}

Scientific hypothesis generation and idea discovery aim to computationally identify unexplored research opportunities, formulate theoretically grounded hypotheses, and reduce the cognitive burden of researchers during the early stages of scientific discovery, thereby accelerating the transition from conceptual exploration to empirical investigation~\cite{1jacobsson2026ai,2bazgir2025agentichypothesis,5alkan2025survey}. According to the knowledge source, reasoning paradigm, and collaboration mechanism, existing approaches can be broadly categorized into three types: single-model parametric generation methods, literature-constrained retrieval-augmented generation methods, and multi-agent discussion-based generation methods.

(1) Single-model parametric generation methods, which directly leverage the internal knowledge of large language models for scientific idea generation. Although efficient and flexible, these methods are prone to hallucination and struggle to distinguish novel discoveries from recombinations of existing knowledge~\cite{2311arlt2025towards}. (2) Literature-constrained retrieval-augmented generation methods, which introduce external scientific evidence such as publications, citation networks, and knowledge repositories to enhance factual grounding and research relevance. Representative systems such as ResearchAgent improve the reliability of generated problems through literature retrieval and validation mechanisms, but still lack detailed methodological reasoning and executable experimental planning~\cite{2331yin2025comprehensive,2312yu2025alpharesearch}. (3) Multi-agent discussion-based generation methods, which employ specialized agents to collaboratively propose, critique, and refine research ideas. Recent frameworks such as SciAgents and The AI Scientist further extend automation toward interdisciplinary discovery and end-to-end research workflows~\cite{7ghafarollahi2025sciagents,lu2026aiScientist}. However, these approaches remain limited by potential reasoning bias among agents and insufficient guarantees of scientific validity~\cite{17beel2025evaluating}.

Overall, existing AI4R systems have demonstrated that literature grounding and multi-agent collaboration can effectively improve the coherence, relevance, and automation of scientific idea generation. However, two critical challenges remain unresolved. First, current multi-agent systems often rely on homogeneous model backbones and shared evidence sources, leading to correlated reasoning biases and potential false consensus among agents rather than effective error correction~\cite{baek2025researchagent}. Second, existing frameworks lack a principled mechanism for compiling natural-language hypotheses into structured and executable research plans, leaving a gap between conceptual ideation and empirical validation~\cite{13yang2026moose}. Therefore, developing socially structured hypothesis generation and research plan compilation mechanisms is essential for improving the reliability, executability, and scientific effectiveness of autonomous research systems, which motivates the BLAZE framework proposed in this work.

\subsection{Multi-agent Scientific Reasoning and Social Collaboration}
\label{sec:multi_agent_science}

Multi-agent scientific reasoning aims to leverage collaboration among multiple
agents with complementary capabilities to overcome the limitations of individual
models in complex scientific discovery. 
According to the organization of agent responsibilities and interaction
mechanisms, existing approaches can be broadly categorized into
role-based collaboration, debate-based deliberation, and hierarchical orchestration.

(1) Role-based collaboration improves scientific problem solving by assigning
agents predefined functional responsibilities. Swanson et al. proposed Virtual Lab, which
organizes multiple scientific specialists into an interdisciplinary team for
nanobody design, enabling collaborative reasoning across different domains~\cite{swanson2025virtual}.
Xie et al. developed CASSIA by separating cell-type annotation, evaluation,
and quality-control processes into cooperating agents
~\cite{xie2026cassia}. These methods demonstrate the
effectiveness of functional specialization in complex scientific workflows.
However, their agent roles are usually predefined for specific tasks, making
the collaboration process difficult to adapt beyond the original workflow. (2) Debate-based deliberation improves scientific reasoning by enabling agents to
generate, criticize, and refine competing hypotheses through iterative
interaction. Du et al. introduced multi-agent debate, where agents exchange
arguments and revise solutions to improve reasoning reliability
~\cite{du2023improving}. Gottweis et al. proposed Co-Scientist, which employs
generation, reflection, ranking, evolution, and meta-review agents to refine
scientific hypotheses through structured debate
~\cite{Gottweis_2026}. Jiang et al. further developed AgenticSciML, where
specialized agents collaboratively propose and refine scientific machine
learning solutions~\cite{jiang2026agenticsciml}. However, debate-based
deliberation mainly focuses on short-term reasoning improvement and lacks
persistent collaboration and evidence-grounded disagreement resolution. (3) Hierarchical orchestration organizes scientific workflows through a central
coordinator and specialized workers to enable task decomposition and
coordination. Ghareeb et al. proposed Robin, which integrates literature
analysis, experimental planning, and laboratory feedback for computational and
physical experiment coordination~\cite{ghareeb2026multiagent}. Boiko et al.
introduced Coscientist, which combines retrieval, planning, code execution,
and experimental control to automate scientific workflows
~\cite{boiko2023autonomous}. Mandal et al. further explored instrument-oriented
agent systems for coordinating laboratory operations and improving
experimental automation~\cite{mandal2025evaluating}. However, existing
hierarchical systems mainly focus on temporary workflow execution and lack
persistent collaboration mechanisms.

Overall, existing multi-agent scientific systems have advanced expertise
coordination, hypothesis generation, and automated experimentation.
Nevertheless, they still lack persistent roles, evidence-grounded conflict
resolution, long-term memory, and clear responsibility attribution.
Therefore, future frameworks should support continuous collaboration and
accountable knowledge accumulation, enabling scientific reasoning to evolve
beyond short-term task execution.
\renewcommand{\arraystretch}{1.35}
\subsection{Automated Experimentation}

Artificial Intelligence for Research (AI4R) aims to reduce the burden of code execution and benchmark evaluation by transforming language interactions into executable scientific tasks and automated workflows. Existing computational research automation can be broadly divided into code-space engineering exploration agents and end-to-end full-lifecycle scientific workflow systems and evaluation frameworks.

(1) Code-space engineering exploration agents~\cite{jiang2025aide, romera2024mathematical, novikov2025alphaevolve} iteratively modify and execute programs under automatic evaluators, and are commonly assessed on MLAgentBench~\cite{huang2023mlagentbench} and MLE-bench~\cite{chan2024mlebench}. They enable efficient code iteration and metric optimization, but task-specific scores alone do not test scientific hypotheses, establish causal attribution, or ensure comparable baselines. (2) End-to-end full-lifecycle scientific workflow systems and evaluation frameworks~\cite{lu2026aiScientist, schmidgall2025agentlab, chen2025mlrbench} coordinate ideation, experimentation, writing, and sometimes automated review. Although they cover a broader research workflow, implementation errors, unreliable self-evaluation, and fabricated or invalid results remain possible; therefore, successful execution or fluent drafts should not be treated as scientific validity without traceable evidence and human verification.

Overall, existing AI research systems can execute code, improve benchmark metrics, and generate research reports, but these capabilities do not by themselves establish scientific validity. Limitations in evidence quality, claim traceability, and human oversight remain important, while LLM feedback and LLM-as-a-judge systems also face reliability and bias concerns~\cite{liang2024llmfeedback,gu2024llmjudge}. BLAZE addresses these gaps by decoupling the knowledge base, \tool{id-pipeline}, and AutoPaper, and by representing ideas as an experiment graph rather than isolated scripts. Its bounded processes, including \tool{auto-baseline}, \tool{idea-validation}, \tool{benchmark-optimize} (and its \tool{-loose} variant), and \tool{review-lite}, link computational work to hypotheses, baselines, and reviewable evidence, while retaining human authors' final responsibility for scientific claims.
\subsection{Automated Peer Review and Human-in-the-Loop Evaluation}

Automated peer review leverages large language models to evaluate research ideas, experiments, and manuscripts, aiming to improve the efficiency and consistency of scientific assessment. Based on the level of automation and human involvement, existing approaches can be categorized into fully automated review, multi-agent review, and
human-in-the-loop evaluation.

(1)Fully automated review methods use a single LLM as an independent evaluator to analyze manuscripts, generate feedback, and assess scientific quality without direct human intervention. Liang et al. investigated the capability of GPT-4 as an automated reviewer by generating feedback on research papers and analyzing its alignment with human reviews, revealing both the potential and limitations of LLM-based evaluation~\cite{liang2024llmfeedback}. Robertson conducted a pilot study on GPT-4-assisted peer review, showing that a single LLM reviewer can provide useful suggestions for improving manuscripts while still struggling with complex expert-level judgment
\cite{robertson2023gpt4slightlyhelpfulpeerreview}. The main advantage of fully automated review lies in its efficiency and scalability, but it remains limited by inconsistent evaluation reliability and insufficient capability for complex scientific judgment. (2)Multi-agent review methods further improve evaluation diversity by assigning different agents complementary reviewing roles. Lu et al. proposed The AI Scientist, which integrates automated reviewers into an end-to-end scientific discovery framework to evaluate generated ideas, experiments, and manuscripts through iterative review feedback~\cite{lu2026aiScientist}. Chen et al. introduced MLR-Bench, which develops MLR-Judge to evaluate AI research agents by using structured assessment criteria and automated review mechanisms~\cite{chen2025mlrbench}. Gottweis et al. proposed Co-Scientist, which employs multiple specialized agents including reflection, ranking, and meta-review agents to iteratively evaluate and refine scientific hypotheses~\cite{Gottweis_2026}. However, reviewers based on similar foundation models may inherit shared
biases, resulting in correlated errors. (3)Human-in-the-loop evaluation combines LLM-based feedback with expert judgment, aiming to improve review efficiency while preserving human responsibility for scientific decisions. Liang et al. analyzed GPT-4-generated feedback on research papers and found partial agreement between LLM comments and human reviews, while revealing limitations in automated evaluation~\cite{liang2024llmfeedback}. Thakkar et al. conducted a large-scale randomized study of LLM feedback in peer review, demonstrating that
AI-assisted suggestions can improve the clarity and actionability of human review reports~\cite{thakkar2026llmfeedback}. The discussion on peer review in the AI era further emphasized that LLMs should serve as supportive tools rather than replace expert judgment, highlighting the importance of human oversight and accountability~\cite{peerreview2026artificial}. Overall, these methods show that LLMs can enhance review efficiency and structure, but human expertise remains essential for scientific judgment.

Therefore, future peer-review frameworks require calibrated AI evaluation, independent assessment mechanisms, and well-designed human-AI interaction to achieve reliable and accountable scientific judgment.
\subsection{Robotic Laboratories, Automated Experimentation and Dark Laboratories}
Automated experimentation connects machine reasoning to instruments, materials, and measurements to reduce the physical bottleneck of discovery. Existing systems can be classified by execution mode and feedback: robotic laboratories automate predefined operations, self driving laboratories use observations to select later experiments, dark laboratories target sustained operation with fault handling, and agentic laboratory systems translate scientific goals into plans and tool calls. These approaches form an autonomy spectrum from repeatable execution to adaptive experimentation and resilient operation rather than equivalent descriptions of one capability.

(1) Robotic laboratories automate synthesis, transport, and characterization; mobile platforms can share existing equipment and combine complementary measurements~\cite{dai2024autonomous}. Their workflows are reproducible but often predefined. (2) Self driving laboratories couple automation with optimization so each result changes the next experiment, as demonstrated by the mobile robotic chemist and A Lab~\cite{hase2019nextgeneration,burger2020mobile,szymanski2023autonomous}. They remain dependent on controlled search spaces. (3) Dark laboratory capabilities add perception and deviation inspection, exemplified by LIRA~\cite{zhou2025localization}, but robust recovery is still immature. (4) Agentic systems such as Coscientist and ChemCrow coordinate retrieval, chemical reasoning, specialist tools, and synthesis~\cite{boiko2023autonomous,bran2024augmenting}, while reliability remains bounded by tool quality and execution verification.

Physical autonomy remains limited by protocol ambiguity, accumulated error, incomplete provenance, and safety risk. Instructions must become explicit quantities, device settings, tolerances, stopping rules, and recovery actions. Measurements, calibration, conditions, sample histories, and failures must remain traceable because one false observation can distort later decisions, while high risk actions require safety controls and human oversight~\cite{seifrid2022autonomous}. BLAZE addresses these gaps through experiment graphs that connect literature grounded plans to controlled execution and return observations, logs, anomalies, and failures to shared memory, keeping human, digital, and physical roles auditable.

\subsection{AI4R Systems and Scientific Organization}
\label{sec:related-gap}

The emerging field of artificial intelligence in scientific research aims to accelerate scientific discovery by supporting the cognitive and execution tasks of research workflows through automation. Existing approaches can be categorized into three types based on their primary role in the research process: literature-based synthesis and creative systems, verifiable computational search and machine learning engineering systems, and end-to-end semi-automated research workflows.

(1) Literature grounded synthesis and ideation systems, such as AutoSurvey and ResearchAgent~\cite{wang2024autosurvey,baek2025researchagent}, retrieve and organize scientific literature to generate surveys, research problems, methods and experimental plans. These systems incorporate retrieval, structured knowledge sources and iterative review. However, they do not generally carry their outputs forward as a shared and auditable project memory that informs subsequent experimentation and scientific claims. (2) Verifiable computational search and machine learning engineering systems, such as AIDE and FunSearch~\cite{jiang2025aide, romera2024mathematical}, use executable evaluation and iterative search to improve code, algorithms and machine learning workflows. Although they maintain search histories and performance records, their objectives are commonly defined over computational metrics rather than explicit links among prior literature, research hypotheses, empirical evidence and manuscript claims. (3) End to end semi autonomous research pipelines, such as The AI Scientist~\cite{lu2026aiScientist}, combine ideation, experimentation, manuscript generation and automated review. Their evidence connections are primarily represented through experiment logs and generated text rather than through an explicit and auditable claim evidence structure. Output quality also varies across tasks and configurations, and human oversight remains part of the workflow.

BLAZE is intended to connect these capabilities within a shared research infrastructure. It maintains knowledge across research phases, supports multiple research roles through discussions tied to a shared evidence state, coordinates closed loop experimentation through automated baseline construction and idea validation, and links scientific communication to explicit claim evidence records. Human researchers retain responsibility for approving research directions, evaluating scientific significance and authorizing consequential decisions.

\section{BLAZE Research System: A Socialized AI4R Architecture}
\label{sec:system}

Existing systems provide partial tool feedback and execution loops, but often lack a unified state representation across the research lifecycle~\cite{lu2026aiScientist}. BLAZE addresses this limitation by modeling research automation as a governed state-transition process rather than unconstrained text generation.

The system has two aligned views. The lifecycle view describes how a project evolves through four macro phases:
Knowledge, Deliberation, Execution, and Validation. The architectural view specifies four shared layers that support
multiple phases rather than mapping rigidly onto them: the knowledge and memory layer, the agent society layer, the
tool and execution layer, and the evaluation and governance layer. Together, these views define the temporal
evolution and structural support of a unified research process.

\subsection{The BLAZE Research Lifecycle}
\label{sec:lifecycle}

\begin{figure}[h]
    \centering
    \includegraphics[width=1\linewidth]{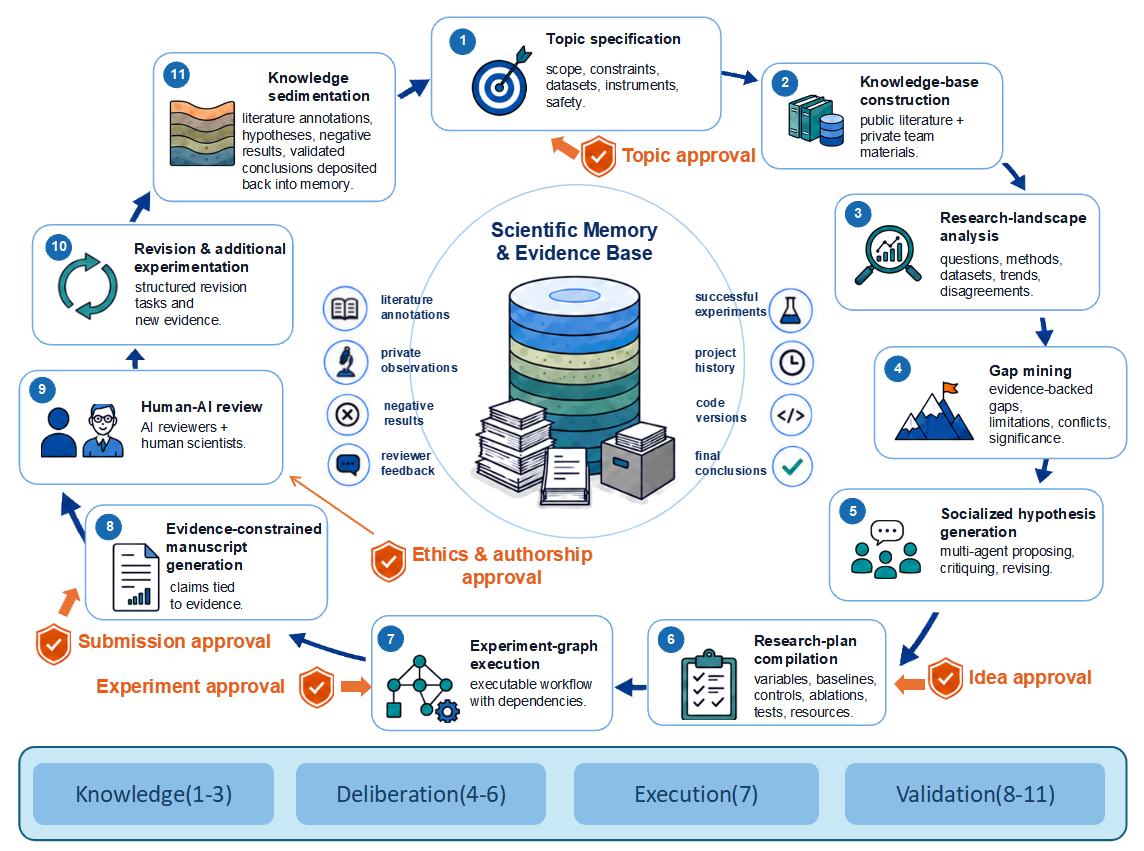}
    \caption{Overview of the BLAZE research lifecycle. The lifecycle comprises four macro phases, Knowledge, Deliberation, Execution and Validation, organized into eleven steps with human approval at key transitions.}
    \label{fig:blaze-lifecycle}
\end{figure}

BLAZE defines the research lifecycle as four macro phases, each instantiated by the fine-grained steps in
Fig.~\ref{fig:blaze-lifecycle}. The \emph{Knowledge} phase transforms an initial human question into a scoped
research context containing disciplinary boundaries, literature evidence, datasets, private project memory, and
field-level disagreements. These fields are active inputs in the current computational workflow. Instrument
interfaces, budget limits, laboratory safety constraints, and physical permissions are reserved schema fields for
future physical experimentation, not fully operational capabilities claimed by the present system.

The \emph{Deliberation} phase includes gap mining, socialized hypothesis generation, and research-plan compilation.
Candidate gaps must carry provenance, supporting evidence, conflicting evidence, and feasibility constraints.
Specialized agents then propose, criticize, and revise hypotheses under falsifiability, methodological plausibility,
resource compatibility, and evidence coverage. Novelty and expected contribution are not objective model-computed
attributes; they are evidence-assisted human judgments. BLAZE can retrieve overlaps, objections, and comparison
evidence, but the final assessment remains with the human researcher or designated PI role.

The \emph{Execution} and \emph{Validation} phases connect approved plans to accountable claims. Execution carries
out the accepted plan and records the resulting artifacts, while Validation checks whether the accumulated evidence
is sufficient for the claims that the manuscript intends to make. Review feedback can reopen earlier assumptions,
require additional evidence, or return the project to a previous phase. Across cycles, validated conclusions,
negative results, unresolved disagreements, and reviewer rationales are sedimented into memory so that later
projects inherit both findings and cautions.

\subsection{Phase-Aligned Architecture and Research Object}
\label{sec:four-layer-architecture}

The four-layer architecture provides shared structural support for the lifecycle rather than a strict one-to-one
mapping from layers to phases. The knowledge and memory layer is most visible during Knowledge construction, but
also serves Validation through citation checks, claim-evidence linking, and reviewer-response memory~\cite{wilkinson2016fair}.
The agent society layer is central to Deliberation, yet also participates in Execution debugging and Validation
review. The tool layer materializes plans as code, notebooks, containers, statistics, figures, documents, or future
instrument calls, while evaluation and governance constrain both Execution and Validation.

To make the correspondence precise, BLAZE represents each project as a \emph{Research Object} \[
\mathcal{R}=\langle q, H, P, G, E, C, V, A, S, \Pi\rangle . \] Here \(q\) is the research question; \(H\) is a set
of hypotheses; \(P\) is a candidate or approved plan; \(G=(N,\mathcal{D})\) is an experiment graph with experiment
nodes \(N\) and dependencies \(\mathcal{D}\); \(E\) is the evidence set; \(C\) is the claim set; \(V\) contains
reviews and revision tasks; \(A\) records human approvals; \(S\) is the project state; and \(\Pi\) stores
provenance, permissions, risk labels, and policy constraints~\cite{moreau2013prov}.

System execution is defined as guarded transitions \(T_i:\mathcal{R}_t\rightarrow\mathcal{R}_{t+1}\). A transition
is valid only if it preserves provenance, updates the relevant fields of \(\mathcal{R}\), and satisfies the quality
gate of the current phase. Knowledge transitions update \(q\), \(\Pi\), and contextual evidence; Deliberation
transitions update \(H\), \(P\), and objections; Execution transitions update \(G\) and \(E\); Validation
transitions update \(C\), \(V\), \(A\), and sedimented memory. This unified state representation prevents
unsupported claims or unapproved actions from moving silently across phase boundaries.

\subsection{Agent Society: From a Single Model to a Research Community}
\label{sec:agent-society}

BLAZE treats multi-agent research as an operational protocol rather than a role-playing prompt~\cite{wu2023autogen}.
Each agent is
defined as \(a=\langle r,\gamma,\kappa,I,O,\Omega,\tau\rangle\), where \(r\) denotes the research role, \(\gamma\)
the activation condition, \(\kappa\) the caller, \(I\) the input schema, \(O\) the output schema, \(\Omega\) the
permitted tool set, and \(\tau\) the stopping rule~\cite{hong2024metagpt}. This definition is needed because role
descriptions alone do not determine activation, authority, evaluation, or termination.

Agent activation is event-driven over the state \(S\) of the Research Object. The PI Agent is called at project
start, phase transitions, and major conflicts; the Literature Agent during Knowledge construction and novelty
checking; the Methodologist and Skeptic Agents during Deliberation; the Experiment and Statistician Agents during
Execution; and the Writer, Reviewer, and Ethics Agents during Validation. The scheduler selects the minimal
predefined set of roles required by the pending transition and passes typed slices of \(\mathcal{R}\). The rule is
a deterministic mapping from transition type, risk label, and missing evidence category to a role set.

Conflict resolution follows a propose-challenge-revise-adjudicate protocol. Critiques must identify the
challenged hypothesis, plan element, evidence item, or claim, and should cite counterevidence or propose a
discriminating experiment. In pseudocode, the transition is \(\mathrm{propose}(x,\mathcal{R})\!\rightarrow\!\mathrm{challenge}(o,E,\Pi)\!\rightarrow\!\mathrm{revise}(o,c)\!\rightarrow\!\mathrm{adjudicate}(o',E,\Pi,A)\).
When evidence, methodological validity, and governance constraints conflict, governance constraints have veto
authority; otherwise, the PI or human controller adjudicates. An agent stops when its schema is complete, its budget
is exhausted, a quality gate decides the transition, or the PI/human controller terminates the task.

\subsection{Human Governance and Intervention Points}
\label{sec:human-governance}

BLAZE is designed as a human-AI collaborative research infrastructure rather than an autonomous submission system~\cite{wickramasinghe2020trustworthy,marusic2010singapore}. Human governance is therefore embedded in the Research Object as part of the research process itself. Five checkpoints cover topic selection, idea formation, experimentation, submission, and ethics and authorship, ensuring that consequential scientific decisions remain under human authority.

Each checkpoint addresses the principal risks of its phase. Topic approval assesses scientific relevance, feasibility and scope. Idea approval examines novelty, evidence and expected contribution. Experiment approval reviews computational cost, data sensitivity and resource use. During Validation, submission and ethics approvals assess claims, evidence, reproducibility, presentation, attribution and responsible disclosure before external communication.

Human decisions are recorded as structured updates to (A). Each record includes the decision owner, timestamp, rationale, scope, object version and artifact hash, allowing later audits to reconstruct what was approved and on which evidence. BLAZE may accelerate search, deliberation, experimentation and drafting, but responsibility for research framing, high-risk actions, dissemination and authorship remains human.

\section{Knowledge Base}
\label{sec:knowledge-base}

The knowledge base is the scientific memory layer of BLAZE Research. Its central function is to preserve research continuity: a downstream workflow should be able to recover which papers were used, which evidence supported a claim, which version of the corpus was queried, and whether the evidence came from public literature or a user's private materials. We therefore do not treat the knowledge base as a single retrieval index. Instead, it is a versioned and provenance-aware substrate that connects papers, full-text chunks, document assets, structured scientific fields, citation topology, author-institution records, review traces, and evidence-linked knowledge graphs.

This design supports continuity in three ways. First, every reusable knowledge object is tied to provenance, including source records, paper identifiers, chunk or asset identifiers, and build versions. Second, public scientific memory and user-specific materials are separated at service and storage boundaries, so that private libraries can be updated, retrieved, or deleted without being mixed with the shared corpus. Third, downstream modules consume evidence snapshots rather than untracked prompt context. As a result, domain dashboards, trend analysis, peer evaluation, evidence-graph inspection, survey writing, and opportunity discovery all operate over a shared but inspectable memory layer.

\subsection{Service Realization}

\begin{figure}[h]
\centering
\includegraphics[width=0.8\linewidth]{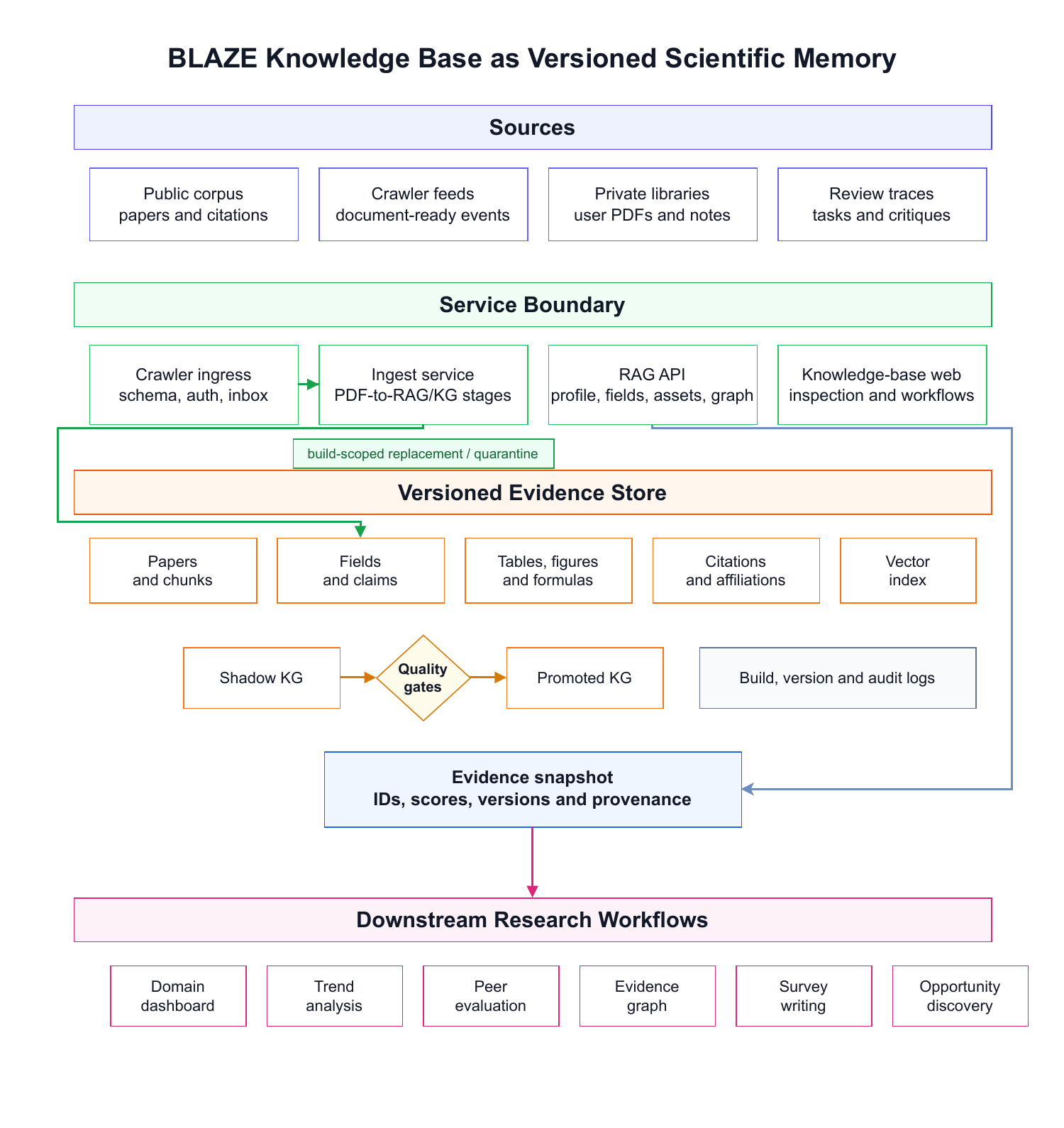}
\caption{Knowledge-layer architecture of BLAZE Research.}
\label{fig:knowledge-layer}
\end{figure}

The memory layer is implemented through four service roles, as summarized in Fig.~\ref{fig:knowledge-layer}. A crawler ingress service validates and queues document-ready events, while an ingestion service manages PDF registration, integrity checks, and the staged PDF-to-RAG/KG pipeline. A RAG API provides retrieval over paper profiles, structured metadata, document assets, graph relations, and evidence packs. A knowledge-base interface supports human inspection and workflow interaction. Auxiliary services, including embedding and staging APIs, function as data-plane dependencies rather than independent knowledge stores.

The ingestion state machine turns a paper into reusable scientific memory through ordered stages, including PDF parsing, base metadata writing, LLM-based field extraction, structured-evidence construction, RAG chunking, vector indexing, citation extraction and linking, KG construction, evidence linking, audit, and completion. Reliability controls are scoped by paper and build identifiers. A new or corrected paper revision replaces only the affected paper/build scope, while existing RAG rows, vector rows, and derived evidence from other builds are protected by build-ownership checks. Failed or suspicious builds can be quarantined or withdrawn so that retrieval clients do not silently consume stale or inconsistent records.

\subsection{Database Coverage and Release-Level Audits}

As of July 2026, the BLAZE knowledge base contains 158,338 papers across 13 venue categories: 12 curated AI conference venue categories and one MIXED category. The 12 curated conference categories account for 157,984 papers, while the remaining 354 papers are assigned to the MIXED category and are excluded from venue-level analyses. At the database level, BLAZE records 90,838 authors, 40,655 institutions, 733,026 paper-author links, 538,785 paper-institution links, 6.06 million paper-citation rows, 9.54 million RAG chunks, 584,997 document assets, and 3.94 million structured field-evidence rows. It further contains 2.35 million extracted claims, 625,034 baseline records, 328,451 dataset records, 313,469 metric records, and a shadow knowledge graph with 3.45 million entities, 6.18 million relations, and 3.57 million relation-evidence links. These statistics describe the current scale and coverage of the BLAZE database rather than the scientific reliability of the extracted content. Data quality is evaluated separately through release-level audits, including duplicate-detection accuracy, field-extraction accuracy, claim-extraction precision, entity-resolution accuracy, evidence-link accuracy, and stale-record rate.

The knowledge layer also records governance metadata. Public records retain source and license information where available. User-private materials are stored and retrieved under user-scoped access boundaries. Deletion requests trigger withdrawal or purge procedures over text chunks, document assets, structured fields, graph edges, and vector-index entries. Each build records a data version, schema version, build identifier, and retrieval-index version, allowing downstream modules to cite not only a paper or chunk but also the knowledge snapshot from which the evidence was retrieved. Updates are event-driven for crawler and PDF-ingestion flows, while retrieval-facing releases are frozen into periodic snapshots for reproducible evaluation and writing.

The shadow KG is a staging and validation graph rather than the final curated KG. Newly extracted entities, aliases, relations, and relation-evidence links are first written into the shadow KG, where provenance, entity resolution, relation conflicts, and evidence support can be inspected. Only records that pass quality gates, version checks, and optional human review are promoted into the stable KG used by long-lived downstream workflows. This separation lets the system absorb new literature quickly without treating every machine-extracted relation as authoritative knowledge.

\begin{table}[!htbp]
\centering
\small
\caption{Downstream evaluation interfaces supported by the knowledge layer.}
\label{tab:knowledge-evaluation}
\begin{tabular}{p{0.22\linewidth}p{0.4\linewidth}p{0.3\linewidth}}
\hline
\textbf{Downstream workflow} & \textbf{Knowledge support} & \textbf{Evaluation signals} \\
\hline
Domain dashboard &
Papers, venues, topics, methods, datasets, authors, institutions, and citation neighborhoods. &
Corpus coverage, dashboard freshness, missing-field rate, and entity-resolution accuracy. \\
Trend analysis &
Time-stamped topics, methods, datasets, metrics, citation relations, and author-institution activity. &
Update latency, stale-record rate, trend stability, and held-out temporal agreement. \\
Peer evaluation &
Reviewer records, review-derived tasks, and human-AI critique traces across revision cycles. &
Human-AI agreement, issue-resolution rate, critique usefulness, and trace completeness. \\
Evidence graph &
Claims, methods, datasets, baselines, metrics, tables, figures, formulas, and KG relations linked to chunks or assets. &
Evidence-link precision, claim-evidence coverage, unsupported-claim rate, and stale-link rate. \\
Survey writing &
Citation topology, evidence packs, structured fields, RAG chunks, assets, and graph relations. &
Citation accuracy, evidence-support coverage, hallucinated-citation rate, and human editing burden. \\
Interdisciplinary discovery &
Cross-field concepts, methods, datasets, metrics, and validation constraints. &
Expert-rated novelty, feasibility, cross-field evidence support, and critique survival rate. \\
\hline
\end{tabular}
\end{table}

Table~\ref{tab:knowledge-evaluation} maps this substrate to downstream research workflows and makes the corresponding evaluation signals explicit. This prevents the knowledge layer from being assessed as a product feature list alone; each supported workflow is tied to measurable coverage, freshness, accuracy, or human-review outcomes.

\subsection{Auditable Retrieval Interface}

The knowledge layer is accessed through retrieval APIs rather than direct database reads. \tool{/retrieve/profile} combines full-text retrieval, BM25-style reranking, and optional dense retrieval for paper-level evidence. \tool{/retrieve/structured} retrieves task, dataset, baseline, metric, and claim fields together with chunk-level support. \tool{/retrieve/assets} returns table, figure, formula, and equation evidence. \tool{/retrieve/graph} retrieves KG relations with attached evidence. \tool{/answer/auto} routes a query to profile or structured retrieval according to query intent, while \tool{/related-work/evidence-pack}, \tool{/related-work/topology}, and \tool{/related-work/draft} assemble citation-backed evidence before drafting.

For auditability, a retrieval response is treated as an evidence snapshot. It exposes the query, rewritten query when applicable, retrieved paper IDs, chunk IDs, asset IDs, field hits, graph relations, scores, retrieval profile, and source metadata. In production workflows, this snapshot should be paired with run-level records that include schema and index versions, caller, timestamp, downstream use, and the selected or ignored evidence. Idea discovery and AutoPaper therefore consume inspectable evidence bundles rather than ungrounded model context. The value of the knowledge base is not only that it stores papers, but that it gives domain dashboards, gap mining, multi-agent discussion, evidence-constrained writing, and review loops a shared, versioned, and auditable scientific memory.

\FloatBarrier

\section{Socialized Hypothesis Generation}
\label{sec:socialized-hypothesis-generation}

Socialized hypothesis generation is a central component of BLAZE. Hypothesis formation is an iterative process constrained by prior evidence, criticism, feasibility, and experimental design.
Research exemplified by AI Scientist has linked idea generation, experimental iteration, and paper writing into an end-to-end process~\cite{lu2026aiScientist}; Co-Scientist, meanwhile, uses specialized agents, including generation, reflection, ranking, evolution, and meta-review, to continuously optimize candidate hypotheses through scientific debate and iterative feedback~\cite{Gottweis_2026}. While these efforts validate the feasibility of automated research workflows, there remains a lack of systematic organization regarding how candidate ideas evolve amid continuous scrutiny from diverse academic roles. To address this, BLAZE systematizes the process of hypothesis generation and discussion, enabling multiple scholar-profile agents to repeatedly discuss, challenge, refine, and consolidate a candidate idea. This produces a candidate hypothesis with explicit objections, evidence boundaries, and an executable evaluation plan.

\subsection{Literature-grounded Gap Mining}

In BLAZE, hypothesis generation begins with the rigorous identification of research gaps, rather than unconstrained topic generation. The knowledge base constructed in Section~\ref{sec:knowledge-base} serves as a map of domain evidence, systematically organizing scientific claims, research methods, datasets, evaluation protocols, stated limitations, public peer-review exchanges, author responses, reproducible results, and negative results. The literature agent organizes these materials into a literature matrix with fields including problem, assumption, method, dataset, baseline, metric, limitation, evidence, and source. This explicit schema supports the identification of research gaps that genuinely require explanation, validation, or redesign.

\begin{table}[h]
    \centering
    \small
    \renewcommand{\arraystretch}{1.25}
    \caption{Structured Gap Record used during literature-grounded gap mining.}
    \label{tab:gap-record}
    \begin{tabular}{p{0.20\linewidth} p{0.42\linewidth} p{0.28\linewidth}}
        \hline
        \textbf{Field} & \textbf{Description} & \textbf{Screening role} \\
        \hline
        Gap type & Includes unresolved problems, methodological bottlenecks, data gaps, evaluation gaps, interdisciplinary transfer opportunities, or persistent peer-review or replication doubts & Prevents vague topics from being treated as research gaps \\
        Supporting evidence & Independent evidence indicating that the gap exists or remains insufficiently addressed & Anchors the research gap in a clear evidence base \\
        Counterevidence & Findings or arguments that may weaken, resolve, or contradict the proposed gap & Enables reverse checking of the evidence \\
        \hline
    \end{tabular}
\end{table}

The search scope for gaps includes unresolved problems, recurring methodological bottlenecks, data gaps, inadequate evaluation criteria, opportunities for interdisciplinary transfer, and persistent doubts arising during peer review or replication; these categories also constitute the gap types obtained from the search. At the same time, a visible research gap is scientifically meaningful only when it is supported by sufficient evidence. Therefore, for candidate research gaps, BLAZE constructs a structured Gap Record, as shown in Table~\ref{tab:gap-record}. A candidate research gap is only considered for further discussion when its type is clearly specified, supporting evidence is sufficient, and counterevidence has been thoroughly examined. This record provides researchers with an evaluation basis while also providing an initial basis for subsequent idea discussion and evolution.

\subsection{Multi-agent Scientific Debate}

\begin{figure}[h]
    \centering
    \includegraphics[width=\linewidth]{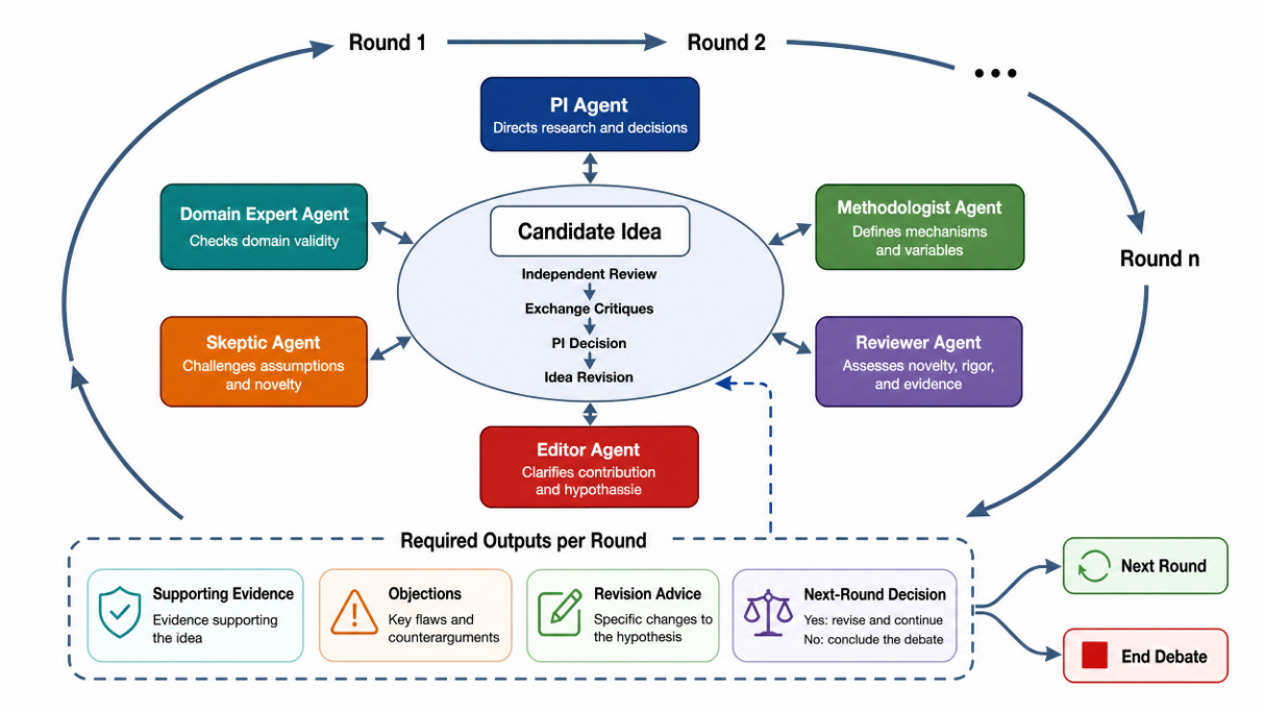}
    \caption{BLAZE brings together PIs, domain experts, methodologists, skeptics, reviewers and editors to evaluate each candidate idea through iterative scientific discussion. Each round records supporting evidence, counterarguments, revision proposals and the decision to advance, revise or reject the idea.}
    \label{fig:multi-agent-scientific-debate}
\end{figure}

Once a potential research gap is identified, BLAZE organizes PI, domain expert, methodologist, skeptic, reviewer, and editor agents to discuss the same concept (Fig. ~\ref{fig:multi-agent-scientific-debate}). These agents assume distinct academic roles rather than acting as interchangeable speakers. The PI agent maintains the research direction and sets decision priorities; the domain expert agent evaluates the concept’s validity within a specific field; the methodologist agent translates the proposition into mechanisms of action, research variables, and alternative explanations; and the skeptic agent identifies hidden assumptions, confounding factors, and overstated claims of innovation. The reviewer agent assesses innovation, rigor, and sufficiency of evidence based on peer review criteria, while the editor agent determines whether the scientific contribution is clear and whether the research has broader scientific significance. This role structure is not intended to increase the number of speakers, but to make conflicts of judgment explicit through a transparent and verifiable division of responsibilities, thereby forming diverse evaluative perspectives and improving hypothesis quality in terms of academic rigor, novelty, and practicality.

While preserving the socialized division of academic roles, BLAZE adds three design constraints. First, persistent roles keep each agent's academic responsibility stable across discussion rounds, so critiques and revisions can continue across rounds and be traced back to a consistent source. Second, evidence-bound objections require challenges to be grounded in literature, data, methods, or experimental records rather than unsupported preference. Third, experimental adjudication converts unresolved disagreements into observable variables, control conditions, evaluation metrics, or follow-up experiments. Together, these commitments produce diverse evaluative perspectives while keeping the discussion traceable and actionable.

Specifically, each role first independently evaluates the candidate concept, articulating arguments for or against specific claims and grounding them in literature, data, methods, or existing experimental records. This process preserves the differences in judgment among different roles while enabling grounded argumentation and exchange of opinions. In each round of discussion, the skeptic, reviewer, and editor agents assume the role of the opposing side and, based on the grounded evidence obtained so far, identify hidden assumptions, alternative explanations, or boundary conditions that could refute the current claims regarding the candidate concept. If a disagreement cannot be resolved based solely on existing evidence, the experimental resolution principle described above is applied. The unresolved disagreement is translated by the Methodologist and Experiment Agents into observable variables, controls, and discriminating experiments. The process proceeds only when supporting evidence, counterarguments, revision suggestions, and next steps have all been documented, and when a new round of discussion is expected to generate new evidence or distinguishable questions. Otherwise, the concept is returned for revision or rejected. The process also terminates when the maximum number of discussion rounds is reached, the discussion budget is exhausted, or a human supervisor stops the discussion. Through multiple rounds of discussion, the final output is not consensus itself, but a critically examined candidate hypothesis with clearly defined evidential boundaries.

\subsection{Hypothesis Archive and Genealogy}

The generation of socialized hypotheses requires a persistent scientific record. Therefore, BLAZE not only retains the output of the latest round of discussions but also records each proposal as a Research Object with version history. Within this Research Object, the Hypothesis Archive serves as the hypothesis-specific substructure that records hypothesis-related content. It preserves the title, research question, hypotheses, draft methodology, experimental plan and its assumptions, risk profile, scoring, discussion history, evidence links, rejected alternatives, and a reference to the lifecycle status maintained by the parent Research Object. It records not only what the system proposes but also how proposals evolve and why certain claims or branches were abandoned. The Hypothesis Archive also preserves failures that hold informational value, revealing whether research questions are saturated, lack data support, or have unfeasible designs. As new data, methods, or resources emerge, these records can be revisited, transforming fleeting discussions into cumulative, reusable scientific assets.

In the process of continuous optimization, a corresponding idea generates an evolutionary lineage that is maintained as a versioned iteration process. The evolutionary lineage can be advanced and constructed according to researchers' requirements. The V1-to-V5 sequence is used as an illustrative lineage. This lineage records development from the initial V1 concept, through the V2 formulation constrained by the literature and the V3 revision addressing scientific objections, to the V4 executable experimental plan and finally to V5, which specifies a validated research plan and claim boundary. Each version transition records the parent version, evidence delta, change rationale, unresolved objections, and approval status. At the same time, each transition is linked to a specific trigger, such as conflicts regarding innovation, methodological flaws, unavailability of datasets, ethical issues, or simulated peer review comments. Consequently, researchers can not only trace the specific reasons behind changes in the concept but also identify resolved and unresolved controversies, thereby informing subsequent projects or experimental planning.

To complement the illustrative lineage above, the following illustrative case shows how version transitions proceed. BLAZE records the evolution of a hypothesis on target detection under varying lighting conditions. V1 considered lighting-aware enhancement or attention, but literature analysis showed overlap with enhancement-before-detection methods. V2 narrowed the question to how lighting stress states relate to detection failure modes. After reviewers noted that a unified enhancement strategy may not adapt to low exposure, glare, and saturation, V3 shifted to detector-side state-aware routing with selective recovery and conditional calibration. V4 then converted the idea into a testable plan using multi-illumination data partitions, baseline comparisons, and metrics such as State AP, Small Object AP, and inference latency. V5 specified a validated research plan and claim boundary centered on illumination state modeling and illumination-aware routing, while retaining the conclusion that “further empirical validation is required.” This illustrative case shows the lineage construction process and how it records key changes, objections, and validation boundaries to support later review and planning.

\subsection{Research Plan Synthesizer}

A mature candidate hypothesis must eventually leave the deliberative space and become a plan that researchers and experimental systems can inspect, execute, and revise. The BLAZE Research Plan Synthesizer transforms the candidate idea into a structured specification containing the research question, falsifiable hypothesis, method, dataset or corpus, baselines, metrics, ablations, risks, expected outcomes, and manuscript storyline. Its purpose is not to produce a polished proposal template, but to verify that a complete chain exists from scientific claim to observable evidence before quality-gate selection.

The Methodologist Agent specifies what must be observed, compared, measured, or manipulated for the hypothesis to be tested. The Experiment Agent checks the availability of data, tools, interfaces, runtime conditions, and logging mechanisms. The Statistician Agent aligns the proposed evidence with uncertainty analysis, significance tests, confidence intervals, sensitivity analyses, or robustness procedures. Because these commitments are made before results are known, the synthesizer reduces opportunities for retrospective justification and exposes designs in which the selected metrics cannot distinguish support from refutation.

\subsection{Quality Gates for Hypothesis Selection}

Before candidate concepts enter the experimental system, BLAZE evaluates them across six dimensions: novelty, significance, falsifiability, feasibility, evidence readiness, and ethical and operational risk. These dimensions respectively assess whether the hypothesis differs from prior work, addresses an important problem, can be exposed to potential refutation, can be implemented under available constraints, has enough prior evidence to justify testing, and can be pursued under acceptable ethical and operational conditions. These gates are designed to screen out under-specified, untestable, poorly grounded, or unsafe hypotheses before experimental resources are committed.

Scores for each dimension are assigned independently by agents responsible for the corresponding academic duties. The Literature Agent, Domain Expert Agent, and Reviewer Agent focus on novelty, significance, and evidence readiness; the Methodologist Agent and Statistician Agent assess falsifiability; the Experiment Agent assesses feasibility; the Skeptic Agent conducts cross-checks across all dimensions; and the Ethics Agent is responsible for ethical and operational risk assessment. The scoring process simultaneously records the opinions of each role and the sources of evidence, ensuring that gatekeeping results can be traced back to specific evaluators. Table ~\ref{tab:hypothesis-quality-gates} further outlines the questions to be evaluated for each dimension, the criteria for judgment, and the procedures to be followed in the event of failure.
\begin{table}[h]
    \centering
    \small
    \renewcommand{\arraystretch}{1.25}
    \caption{The six-dimensional quality gate system for hypothesis screening. Each gate is associated with a specific issue of concern, the sources of evidence used, and procedures for handling low-scoring ideas.}
    \label{tab:hypothesis-quality-gates}
    \begin{tabular}{p{0.14\linewidth} p{0.23\linewidth} p{0.25\linewidth} p{0.25\linewidth}}
        \hline
        \textbf{Gate} & \textbf{Evaluation Question} & \textbf{Evidence basis} & \textbf{Action if Below Threshold} \\
        \hline
        Novelty & Does the hypothesis meaningfully differ from existing work? & Literature matrix, nearest-work comparison and peer assessment & Return to gap mining or sharpen the research distinction \\
        Significance & Does the hypothesis address an important scientific or practical problem? & Field trends, expert assessment and citation networks & Revise the problem framing or application context \\
        Falsifiability & Can the hypothesis be stated so that evidence could support or refute it? & Variables, controls, baselines, metrics and statistical design & Reformulate claims, variables, controls, or evaluation criteria \\
        Feasibility & Can the study be executed with available data, tools, budget and time? & Datasets, toolchain, compute budget and access constraints & Narrow the task, stage the objective, or secure missing resources \\
        Evidence readiness & Is there enough prior grounding to justify testing the hypothesis & Literature matrix, prior experiments, reproducibility records and negative results & Gather additional literature, define a pilot study, or return to gap mining \\
        Ethical and operational risk & Are academic, societal, privacy, safety, or operational risks acceptable and governable? & Ethics checklist, data governance, deployment assumptions and human-approval requirements & Add safeguards, require human approval, restrict scope, or terminate the process \\
        \hline
    \end{tabular}
\end{table}

There are varying degrees of conflict among the six dimensions. A highly novel hypothesis may have fewer directly applicable precedents, reducing evidence readiness; a significant problem may require data, resources, or governance conditions that exceed current feasibility; and a feasible implementation can still be insufficiently falsifiable if its baselines, metrics, or statistical designs cannot distinguish support from refutation. BLAZE therefore adopts non-compensatory gatekeeping and does not use scores from other dimensions or a single overall score to offset obvious deficiencies in any one dimension: high novelty cannot compensate for an unfalsifiable hypothesis, and high significance cannot compensate for unacceptable ethical or operational risk. Instead, it retains these conflicts and transforms them into issues to be addressed through subsequent discussion, supplementary justification, or pilot experiments.

The system supports configurable thresholds for each evaluation dimension. These thresholds can be adjusted based on the research stage, field-specific norms, resource conditions, and research objectives, thereby guiding the direction of the concept’s evolution. Projects emphasizing originality may raise the bar for novelty. Early exploratory projects may allow lower evidence readiness while requiring stronger falsifiability. Resource-constrained or risk-sensitive projects may tighten feasibility and governance constraints. Candidate concepts advance to the experimental phase only after meeting all necessary thresholds; dimensions that fail to meet the criteria are returned to the corresponding stage for revision according to the path outlined in the table. This six-dimensional quality gate supports the scientific quality and experimental maturity of ideas through multi-dimensional assessment and continuous refinement.

\setcounter{section}{5}
\section{Closed-loop Experimentation: From Hypotheses to Evidence}
\label{sec:closed-loop-experimentation}

Closed-loop experimentation is the epistemic boundary between a research assistant that produces plausible text and an \aifourr{} system that produces inspectable evidence. The central problem is not simply to execute code, but to preserve the meaning of a test while a research plan is translated into data operations, implementations, optimization, repeated execution, statistical analysis, and finally a manuscript claim. Recent machine-learning agent benchmarks treat execution as a research capability~\cite{huang2023mlagentbench,chan2024mlebench}; a scientific workflow must additionally make its protocol, provenance, and inferential limits recoverable.

Within BLAZE, the hypothesis-to-evidence transition is governed by three invariants. \emph{Comparability} requires a claimed contrast to inherit the same task definition, split, metric, and baseline contract unless a deviation is declared. \emph{Provenance} keeps each result linked to the code, environment, data, configuration, logs, and decisions that produced it. \emph{Calibration} requires the strength and scope of a conclusion to match the validation actually performed. Together, these invariants turn an isolated score into a reusable scientific object.

The AutoPaper stage sequence and artifact responsibilities provide the operational foundation for BLAZE's closed-loop experimentation workflow. This section turns that foundation into an auditable experiment graph and a shared measurement contract, enabling the current evaluation to examine lifecycle coverage, evidence completeness, and trace-supported execution behavior within a consistent evidentiary framework. Quantitative indicators, including first-run validity, repair effectiveness, time efficiency, and agentic-compute scaling, are calibrated to the provenance and comparison design of the available runs, making present results inspectable while supporting progressively broader controlled evaluation.

\subsection{From an Approved Plan to an Experiment Graph}
\label{sec:experiment-graph}

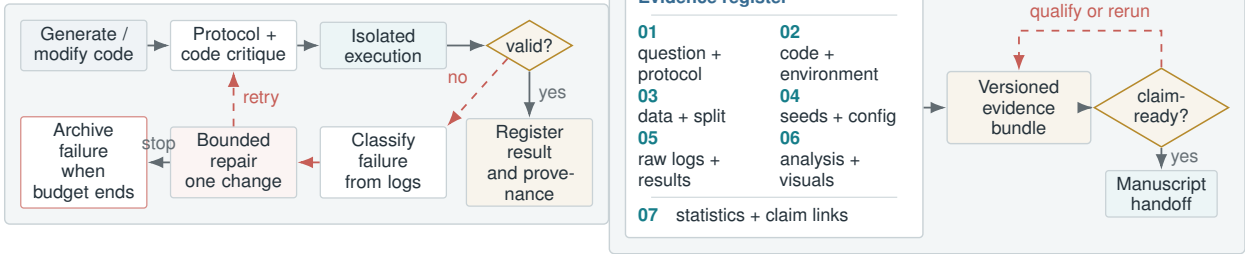
\begin{figure*}[h]
    \centering
    \resizebox{\textwidth}{!}{
\begin{tikzpicture}[
    x=1cm,
    y=1cm,
    font=\sffamily\scriptsize,
    >=Latex,
    stage/.style={
        draw=BaizeLine,
        line width=0.55pt,
        rounded corners=1.5pt,
        minimum height=0.82cm,
        text width=1.85cm,
        align=center,
        inner sep=3.5pt,
        text=BaizeInk,
        fill=white
    },
    compact/.style={
        draw=BaizeLine,
        line width=0.5pt,
        rounded corners=1.2pt,
        minimum height=0.58cm,
        text width=1.55cm,
        align=center,
        inner sep=2.5pt,
        text=BaizeInk,
        fill=white
    },
    registry/.style={
        draw=BaizeLine,
        line width=0.55pt,
        rounded corners=1.5pt,
        align=left,
        inner xsep=5pt,
        inner ysep=4pt,
        text=BaizeInk,
        fill=white
    },
    decision/.style={
        diamond,
        draw=BaizeGold,
        fill=BaizeGold!9,
        line width=0.6pt,
        aspect=1.7,
        inner sep=1.2pt,
        align=center,
        text=BaizeInk
    },
    flow/.style={-Latex, draw=BaizeInk!72, line width=0.7pt},
    branch/.style={-Latex, draw=BaizeBlue!78, line width=0.65pt},
    feedback/.style={-Latex, draw=BaizeCoral, line width=0.7pt, dashed},
    panel/.style={
        draw=BaizeLine,
        fill=BaizeMist,
        line width=0.55pt,
        rounded corners=2pt,
        inner sep=6pt
    },
    paneltitle/.style={font=\sffamily\bfseries\footnotesize, text=BaizeInk}
]

\node[stage, fill=BaizeBlue!8] (handoff) at (0.2,0) {Research brief\\\textcolor{BaizeBlue}{question + protocol}};
\node[stage] (baseline) at (2.65,0) {Baseline\\reproduction};
\node[stage, fill=BaizeTeal!9] (main) at (5.1,0) {Main-method\\validation};
\node[compact] (tune) at (7.55,0.65) {Constrained\\tuning};
\node[compact] (ablate) at (7.55,0) {Ablation};
\node[compact] (robust) at (7.55,-0.65) {Robustness +\\replication};
\node[stage, fill=BaizeGold!9] (aggregate) at (10.0,0) {Protocol-aware\\aggregation};
\node[stage, fill=BaizeCoral!8] (claim) at (12.45,0) {Qualified evidence\\for a scoped claim};
\node[compact, draw=BaizeCoral!70, fill=BaizeCoral!5] (failure) at (5.1,-1.55) {Failure / deviation\\record};

\draw[flow] (handoff) -- node[above, text=BaizeInk!70] {lock} (baseline);
\draw[flow] (baseline) -- node[above, text=BaizeInk!70] {anchor} (main);
\draw[branch] (main.east) -- ++(0.48,0) |- (tune.west);
\draw[branch] (main.east) -- (ablate.west);
\draw[branch] (main.east) -- ++(0.48,0) |- (robust.west);
\draw[branch] (tune.east) -- ++(0.38,0) |- (aggregate.west);
\draw[branch] (ablate.east) -- (aggregate.west);
\draw[branch] (robust.east) -- ++(0.38,0) |- (aggregate.west);
\draw[flow] (aggregate) -- node[above, text=BaizeInk!70] {gate} (claim);
\draw[feedback] (main.south) -- (failure.north);
\draw[feedback] (failure.west) -| node[pos=0.72, left, text=BaizeCoral] {diagnose} (baseline.south);

\begin{scope}[on background layer]
    \node[panel, fit=(handoff)(baseline)(main)(tune)(ablate)(robust)(aggregate)(claim)(failure)] (panelA) {};
\end{scope}
\node[paneltitle, anchor=south west] at ([xshift=1pt,yshift=2pt]panelA.north west) {\textcolor{BaizeTeal}{a}\quad Experiment graph: comparability is inherited along explicit dependencies};

\node[compact, fill=BaizeBlue!8] (generate) at (0.0,-4.15) {Generate /\\modify code};
\node[compact] (critic) at (2.05,-4.15) {Protocol +\\code critique};
\node[compact, fill=BaizeTeal!8] (execute) at (4.1,-4.15) {Isolated\\execution};
\node[decision] (valid) at (6.1,-4.15) {valid?};
\node[compact, fill=BaizeGold!9] (register) at (6.1,-5.75) {Register result\\and provenance};
\node[compact] (diagnose) at (4.1,-5.75) {Classify failure\\from logs};
\node[compact, fill=BaizeCoral!7] (repair) at (2.05,-5.75) {Bounded repair\\one change};
\node[compact, draw=BaizeCoral!70] (archive) at (0.0,-5.75) {Archive failure\\when budget ends};

\draw[flow] (generate) -- (critic);
\draw[flow] (critic) -- (execute);
\draw[flow] (execute) -- (valid);
\draw[flow] (valid) -- node[right, text=BaizeInk!70] {yes} (register);
\draw[feedback] (valid.south west) -- node[above left, text=BaizeCoral] {no} (diagnose.north east);
\draw[feedback] (diagnose) -- (repair);
\draw[feedback] (repair.north) -- node[right, text=BaizeCoral] {retry} (critic.south);
\draw[flow] (repair) -- node[above, text=BaizeInk!70] {stop} (archive);

\begin{scope}[on background layer]
    \node[panel, fit=(generate)(valid)(archive)(register)] (panelB) {};
\end{scope}
\node[paneltitle, anchor=south west] at ([xshift=1pt,yshift=2pt]panelB.north west) {\textcolor{BaizeBlue}{b}\quad Execution and bounded repair};

\node[registry] (evidence) at (9.45,-5.0) {%
    \begingroup
    \setlength{\tabcolsep}{0pt}%
    \renewcommand{\arraystretch}{1.08}%
    \arrayrulecolor{BaizeLine}%
    \begin{tabular}{@{}>{\raggedright\arraybackslash}p{1.78cm}@{\hspace{0.16cm}}>{\raggedright\arraybackslash}p{1.78cm}@{}}
        \multicolumn{2}{@{}l@{}}{\textcolor{BaizeBlue}{\bfseries Evidence register}} \\
        \cmidrule(lr){1-2}
        \textcolor{BaizeTeal}{\bfseries 01}\par question + protocol &
        \textcolor{BaizeTeal}{\bfseries 02}\par code + environment \\[2pt]
        \textcolor{BaizeTeal}{\bfseries 03}\par data + split &
        \textcolor{BaizeTeal}{\bfseries 04}\par seeds + config \\[2pt]
        \textcolor{BaizeTeal}{\bfseries 05}\par raw logs + results &
        \textcolor{BaizeTeal}{\bfseries 06}\par analysis + visuals \\
        \cmidrule(lr){1-2}
        \multicolumn{2}{@{}l@{}}{\textcolor{BaizeTeal}{\bfseries 07}\quad statistics + claim links}
    \end{tabular}%
    \endgroup
};
\node[stage, text width=1.72cm, fill=BaizeGold!9] (bundle) at (12.8,-5.0) {Versioned\\evidence bundle};
\node[decision] (gate) at (14.75,-5.0) {claim-\\ready?};
\node[compact, text width=1.35cm, fill=BaizeTeal!8] (handover) at (14.75,-6.18) {Manuscript\\handoff};

\draw[flow] (evidence.east) -- (bundle.west);
\draw[flow] (bundle) -- (gate);
\draw[flow] (gate) -- node[right, text=BaizeInk!70] {yes} (handover);
\draw[feedback] (gate.north) -- ++(0,0.5) -| node[pos=0.25, above, text=BaizeCoral] {qualify or rerun} (bundle.north);

\begin{scope}[on background layer]
    \node[panel, fit=(evidence)(bundle)(gate)(handover)] (panelC) {};
\end{scope}
\node[paneltitle, anchor=south west] at ([xshift=1pt,yshift=2pt]panelC.north west) {\textcolor{BaizeGold}{c}\quad Evidence assembly and claim gate};

\end{tikzpicture}}
    \caption{Closed-loop protocol and evidence boundary. \textbf{a}, a versioned plan is compiled into an experiment graph whose branches share a baseline and end at a claim gate. \textbf{b}, execution follows a bounded cycle of generation, critique, execution and diagnosis; exhausted repairs are archived as failures rather than concealed by repeated retries. \textbf{c}, manuscript handoff requires a versioned evidence bundle. The figure illustrates system behavior, not measured performance.}
    \label{fig:closed-loop-protocol}
\end{figure*}

An approved plan is the minimum scientific handoff. It fixes the research question, falsifiable hypothesis, method, data and split, baselines and controls, metric direction, planned ablations, statistical analysis, resource budget, safety constraints, and stopping conditions before execution begins. Expected outcomes may motivate the design, but they are not evidence and are never copied into a result record. This boundary prevents an agent from filling an underspecified plan with assumptions that later appear to be experimental facts.

BLAZE compiles the handoff into a typed directed graph
\begin{equation}
    \mathcal{G}_{\mathrm{exp}}=(V,E_{\mathrm{dep}},E_{\mathrm{prov}},E_{\mathrm{claim}}),
\end{equation}
where $V$ contains preparation, baseline, main-method, tuning, ablation, robustness, replication, debugging, and aggregation nodes. Dependency edges $E_{\mathrm{dep}}$ determine execution order; provenance edges $E_{\mathrm{prov}}$ connect a node to the artifacts and decisions from which it was derived; and candidate-claim edges $E_{\mathrm{claim}}$ state which claims a validated node may support or contradict~\cite{moreau2013prov}. Each node stores a stable identifier, plan version, comparison target, dataset and split, metric, configuration, seed, resource ceiling, stopping rule, raw-log location, parsed output, and status. Scientific validation remains a separate gate from process completion.

The AutoPaper stages are explicitly represented within this graph. \tool{auto-baseline} establishes the empirical reference and records reproduction discrepancies; \tool{idea-validation} evaluates research hypotheses; \tool{benchmark-optimize} explores predefined optimization spaces; \tool{benchmark-optimize-loose} supports clearly separated exploratory investigations; and \tool{review-lite} initiates additional analyses, such as ablations and robustness evaluations. Fig.~\ref{fig:closed-loop-protocol} illustrates this closed-loop process. Baseline fidelity serves as a foundation for credible comparison: task definitions, data splits, preprocessing procedures, evaluation metrics, computational budgets, and baseline configurations must follow the approved protocol or be explicitly documented as deviations. Any methodological change that alters the basis of comparison is preserved as a new research branch rather than replacing previous evidence.

\FloatBarrier
\subsection{Code Generation, Execution, and Debugging}
\label{sec:code-debug}

Execution is organized as a separation of responsibilities rather than a single agent call. The Experiment Agent implements one graph node; the Code Critic checks the patch against the method specification and frozen comparison contract; the Executor runs it in the approved environment and captures exit status, streams, resource use, and artifacts; and the Debug Agent diagnoses a failure from the retained trace. The Statistician Agent operates only after output validation, so code repair cannot be used to revise an interpretation that is already being assessed.

Repair is bounded and hypothesis-preserving. Each attempt records its parent, failure class, proposed cause, patch, changed parameters, and outcome; classes include environment, dependency, data, implementation, numerical, resource, and protocol failures. A repair may correct an implementation or environment defect, but it may not silently change the dataset, split, metric, baseline, or objective. Non-running baselines, unstable training, resource exhaustion, null results, and rejected repairs remain distinct records, because collapsing them into one ``failed'' label removes information about both feasibility and scientific validity.

Optimization follows the same integrity rule. Conservative tuning may vary only predeclared repository-supported controls such as learning rate, scheduler, batch size, regularization, threshold, or training budget. Model changes, test-time augmentation, ensembles, or test-set feedback belong to an explicitly labelled exploratory branch and cannot be reported as held-out generalization. The workflow therefore distinguishes an executed score, an automatically repaired run, and a protocol-valid comparison; only the last can support a claim at the stated scope.

\FloatBarrier
\subsection{Data and Tool Integration}
\label{sec:data-tool-integration}

\begin{table}[h]
\caption{Governed connector contract for closed-loop experimentation. ``Required'' denotes the present computational scope; ``Roadmap'' denotes a conditional physical or domain-specific extension.}
\label{tab:tool-interfaces}
\centering
\small
\renewcommand{\arraystretch}{1.16}
\setlength{\tabcolsep}{4pt}
\begin{tabularx}{\textwidth}{@{}>{\raggedright\arraybackslash}p{0.14\textwidth}>{\raggedright\arraybackslash}p{0.19\textwidth}>{\raggedright\arraybackslash}X>{\raggedright\arraybackslash}X>{\raggedright\arraybackslash}p{0.11\textwidth}@{}}
\toprule
\textbf{Interface} & \textbf{Inputs} & \textbf{Identity and record} & \textbf{Gate} & \textbf{Scope} \\
\midrule
Literature and knowledge 
& Papers, DOI/arXiv metadata 
& Source identifier, version/time, extracted field, citation key 
& Verify source; retain stale or conflicting metadata 
& Required \\

Code and execution 
& Repository, notebook, container 
& Commit/content hash, environment lock, invocation, logs, artifacts 
& Rebuild environment; enforce command and resource scope 
& Required \\

Data 
& Dataset or public service 
& Version/checksum, split, preprocessing, license 
& Check permission, leakage, and split fidelity 
& Required \\

Analysis and documents 
& Results, scripts, plotting, \LaTeX 
& Raw input, script version, parameters, generated table/figure 
& Recompute; check labels, units, and metric direction 
& Required \\

Domain and physical 
& Simulator or instrument 
& Tool version, operating envelope, authorization, safety log 
& Owner approval and fail-safe review 
& Roadmap \\

\bottomrule
\end{tabularx}
\end{table}

Closed-loop experimentation crosses systems with different identities, permissions, and failure semantics: literature records, repositories, notebooks, containers, datasets, compute services, plotting and typesetting tools, simulators, and, eventually, physical instruments. BLAZE models each connector as a governed boundary rather than an untyped tool call. The connector contract declares accepted inputs, produced artifacts, identity and version fields, permission scope, license constraints, expected failures, and audit events; the examples in Table~\ref{tab:tool-interfaces} describe required records, not a claim that every interface is already implemented.

The data contract is the main protection against a technically runnable but scientifically ambiguous result. Dataset identity alone is insufficient: split, preprocessing, label map, filtering, augmentation, and any use of validation or test feedback determine what a score means. Immutable versions or checksums make changes detectable; access and license fields prevent an unauthorized experiment. For private inputs, the record stores only the provenance required for an authorized audit and does not expose protected contents to unrelated agents.

The present method covers literature-grounded work, computational experiments, and simulation when the declared connector and audit records are available. Physical laboratory and robotics interfaces require domain-specific safety controls, authorization, and action logs, so they are treated as roadmap capability rather than demonstrated closed-loop performance. This distinction keeps interface breadth separate from evidence of execution depth.

\FloatBarrier
\subsection{Reproducibility and Statistical Validation}
\label{sec:reproducibility}

Reproducibility is an artifact contract, not a declaration. A claim-eligible result carries a versioned evidence bundle with the approved protocol; code and environment versions; data identity, split, and preprocessing; seeds and configurations; complete logs and raw outputs; parsing and statistical scripts; editable figure and table sources; and links to claims that use or contradict the result~\cite{wilkinson2016fair,sandve2013reproducible}. Semantic checks compare parsed metrics with raw outputs, reconcile sample counts, verify units and direction, and regenerate visual material from retained source data.

Statistical validation is claim- and domain-dependent. Before inspecting outcomes, the plan specifies the estimand, independent experimental unit, comparison, stopping rule, repeated-run rule, aggregation method, and uncertainty summary. When hypothesis tests are appropriate, the test, assumptions, significance criterion, and multiple-comparison treatment are also predeclared; effect estimates and uncertainty are reported instead of a threshold alone~\cite{wasserstein2016asa}. If those assumptions do not hold, the workflow retains the empirical distribution or uses a domain-appropriate alternative rather than forcing a generic test.

The evidence bundle distinguishes an \emph{exploratory} run, a \emph{protocol-valid} result, and a \emph{replicated} result. Exploratory runs may support diagnosis but cannot establish a confirmatory claim; protocol-valid results support a scoped observation; replicated results additionally satisfy the declared repeated-run or independent-replication requirement. Failed, null, and inconclusive outcomes remain first-class records. Human approval is mandatory for protocol deviations, sensitive data, high-stakes interpretation, or domain-specific statistical judgement; if the environment cannot be reconstructed, the result is marked non-reproducible.

\FloatBarrier
\subsection{Scaling with Agentic Compute}
\label{sec:agentic-compute}

Agentic compute is the budget for deliberation, execution, repair, review, and human interaction around a scientific task. It is represented as a resource vector rather than as a count of agents:
\begin{equation}
    \mathbf{B}=(n_{\mathrm{delib}},n_{\mathrm{exp}},n_{\mathrm{debug}},n_{\mathrm{review}},t_{\mathrm{human}},C_{\mathrm{total}}),
    \qquad
    C_{\mathrm{total}}=C_{\mathrm{model}}+C_{\mathrm{compute}}+C_{\mathrm{tools}}+C_{\mathrm{human}}.
\end{equation}
The components remain separate because model calls, GPU time, external tools, and expert attention are not interchangeable. More budget may reveal a protocol error or improve replication, but it may also amplify redundant search, test-feedback overfitting, or reviewer disagreement; no monotonic quality trend is assumed.

A valid scaling study compares predeclared budget levels on the same paired task set while holding model family, tool access, data, and evaluation protocol fixed. Stopping conditions and eligible retries are frozen, and adjudicators of scientific quality are blinded to budget where feasible. The analysis reports distributions for each quality endpoint and each cost component with uncertainty, then identifies a cost-quality Pareto frontier instead of collapsing heterogeneous resources into an unsupported ratio. Table~\ref{tab:experiment-metrics} extends these controls into a common measurement contract, fixing the denominator and interpretation boundary for each workflow, evidence, quality, and cost endpoint before cross-budget or cross-system comparison.

\begin{table}[H]
\caption{Measurement contract for closed-loop experimentation and quantitative system comparison. Denominators and reporting rules are fixed before evaluation; this section contains no observed system value.}
\label{tab:experiment-metrics}
\centering
\small
\renewcommand{\arraystretch}{1.16}
\setlength{\tabcolsep}{4pt}
\begin{tabularx}{\textwidth}{@{}>{\raggedright\arraybackslash}p{0.20\textwidth}>{\raggedright\arraybackslash}X>{\raggedright\arraybackslash}X@{}}
\toprule
\textbf{Endpoint} & \textbf{Denominator and computation} & \textbf{Required interpretation} \\
\midrule
First-run validity 
& Eligible runnable nodes passing protocol checks on attempt one / eligible runnable nodes 
& Separate process execution from scientific validity; disclose blocked nodes \\

Bounded debug recovery 
& Failed nodes repaired within the frozen budget without protocol drift / failed nodes approved for repair 
& Report failure class, attempts, time, and changes \\

Experiment completion 
& Nodes reaching a predeclared terminal state / nodes in the frozen plan 
& Separate succeeded, failed, blocked, invalidated, and stopped \\

Lifecycle coverage 
& Planned stages with an admissible artifact / stages in the shared protocol 
& Report stage-level coverage; coverage is not execution success \\

Evidence completeness 
& Required provenance fields present / fields required by the evidence schema 
& Report missing fields and contradictions; do not impute \\

Human involvement 
& Approval, override, and review events / stage events 
& Report event counts and human time with the measurement method \\

Research output quality 
& Blinded rubric score / maximum rubric score 
& Report rubric, assessor count, and agreement; not scientific impact \\

Replication stability 
& Distribution of the predeclared estimate over eligible independent runs 
& Report estimate, uncertainty, failure frequency, run count, and aggregation \\

Claim-evidence coverage 
& Empirical claims linked to admissible evidence / substantive empirical claims 
& Audit both support and contradiction; a link is not sufficiency \\

Cost-quality frontier 
& Paired quality endpoints at each budget with model, compute, tool, and human costs 
& Report marginal changes and dominated settings; avoid uncalibrated ratios \\

\bottomrule
\end{tabularx}
\end{table}

The same contract makes the complementary evaluation tracks in Section~\ref{sec:evaluation} auditable without collapsing them into a single system ranking. Frozen packages are assessed through dimension-level evidence profiles, while shared-task or common-testbed evidence is interpreted within its declared protocol, estimand, and source availability; outcomes obtained under unmatched execution conditions are not aggregated as system performance. Values absent from retained artifacts or public documentation remain ``not reported,'' ``not evaluated,'' or N/A rather than being imputed as zero, and reported, artifact-observed, and independently reproduced evidence are kept distinct. The resulting scores and comparison tables therefore characterize retained research evidence and observable task support under documented conditions; causal attribution or agentic-compute scaling would require matched models, inputs, compute and time budgets, human-intervention policies, and stopping rules.

\FloatBarrier
\subsection{Evidence Qualification and Manuscript Handoff}
\label{sec:evidence-handoff}

The terminal output of closed-loop experimentation is not a folder of favorable scores, but a collection of versioned evidence bundles. Each bundle associates a scoped candidate claim with supporting and contradicting experiment-graph nodes, baseline-fidelity status, statistical summaries, protocol deviations, known limitations, editable figure and table sources, and the relevant human authorization state. Its constituent results may be executable but not protocol-valid, protocol-valid but exploratory, or replicated yet too narrow for a broad claim. At manuscript handoff, the bundle exposes these result records as the typed experiment packets used in Section~\ref{sec:evidence-packets}, where they can be matched with literature and review packets without losing bundle-level provenance.

All evidence bundles are transferred to the evidence-constrained manuscript stage, while the claim-readiness gate determines how their constituent packets may be used. Claim-ready evidence may support scoped empirical statements; failed, inconclusive, or weak evidence is instead routed to limitations, rejected-hypothesis traces, or follow-up actions. The handoff preserves a reversible link to $\mathcal{G}_{\mathrm{exp}}$: a missing control, unresolved uncertainty, or later reviewer objection creates a new approved branch instead of rewriting the original record.

Human researchers retain authority over research purpose, protocol deviations, sensitive-data use, claim strength, authorship, and submission. Agents may organize artifacts, diagnose failures, recompute statistics, and draft bounded revisions, but they may not fabricate evidence, suppress unsuccessful attempts, or convert roadmap interfaces into demonstrated capability. At the current scope, BLAZE can claim a governed mechanism and a falsifiable measurement protocol; stronger claims about reliability, efficiency, or generalization require the paired experiments and retained evidence defined above.

\setcounter{section}{6}

\providecommand{\BLAZE}{BLAZE}
\providecommand{\aifourr}{AI4R}
\providecommand{\tool}[1]{\texttt{#1}}

\section{Evidence-constrained Manuscript Generation}
\label{sec:evidence-constrained-writing}

A manuscript claim must not exceed the evidentiary state of the project. AutoSurvey shows that models can organize literature retrieval, outline construction, section writing, and survey evaluation~\cite{wang2024autosurvey}, and The AI Scientist connects idea generation, code execution, experimental analysis, manuscript drafting, and automated review in an end-to-end research loop~\cite{lu2026aiScientist}. These systems show that research text can be produced at scale. They also expose a harder infrastructure problem: every claim, citation, figure, limitation, and revision must remain connected to the evidence that permits it.

\BLAZE{} therefore defines AutoPaper as an evidence-constrained writing mechanism rather than as a generic paper generator. Its input is not a blank prompt, but a structured record of literature, hypotheses, executable experiments, logs, figures, review comments, and human decisions. Its output is a draft in which claims, citations, figures, limitations, and hedge levels can be traced to explicit evidence packets. This design separates the technical mechanism of writing from the responsible-use norms that govern when a claim may enter a paper.

\begin{figure}[htbp]
\centering
\includegraphics[width=0.65\linewidth]{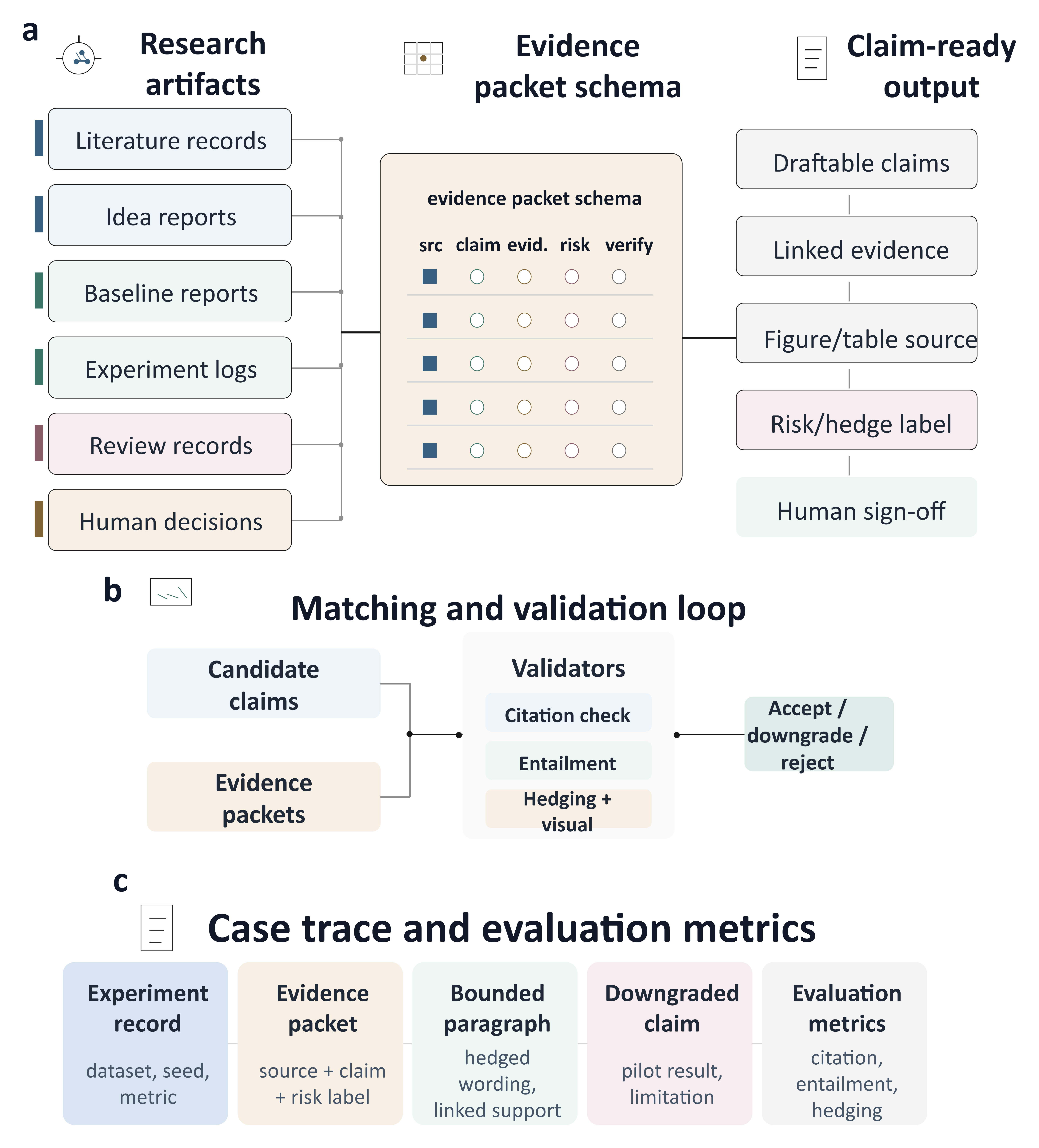}
\caption{\textbf{Evidence-constrained manuscript generation.} \textbf{a,} Literature records, idea reports, baseline reports, experiment logs, review records, and human decisions are compiled into evidence packets with source identifiers, candidate claims, evidence payloads, risk labels, validator states, and human ownership. \textbf{b,} Candidate claims are matched to packets and passed through provenance, citation, entailment, figure/table, and hedging validators before they can enter the draft. \textbf{c,} A case trace shows how an experiment record becomes an evidence packet, a bounded paragraph, a downgraded claim, and a claim-level audit record for citation fidelity, entailment, and hedging.}
\label{fig:evidence-constrained-generation}
\end{figure}

Fig.~\ref{fig:evidence-constrained-generation} summarizes this design. Panel a describes how upstream research artifacts are converted into evidence packets; panel b shows how the matcher and validators constrain writing; panel c illustrates how a manuscript paragraph is accepted, downgraded, or rejected according to evidence strength. The section follows the same order: packet schema, matching algorithm, validator outputs, case trace, and responsibility boundary.

\subsection{Evidence Packets as The Writing Substrate}
\label{sec:evidence-packets}

The first technical problem is representation. If experimental logs, citation notes, figure files, and reviewer comments remain as disconnected documents, a writer model can compose a plausible paragraph whose support is unclear. AutoPaper addresses this problem by converting heterogeneous project artifacts into evidence packets. Each packet is a typed record with a packet identifier, source identifier, artifact type, provenance record, source reliability label, claim scope, evidence payload, evidence strength, required citation, linked figure or table, validator result, claim status, and human owner.

\begin{table}[htbp]
\caption{Core fields of an AutoPaper evidence packet. The packet separates technical provenance from writing permission, so that a fluent claim cannot enter the manuscript without an auditable support record.}
\label{tab:evidence-packet-schema}
\centering
\small
\begin{tabular}{p{0.22\linewidth}p{0.34\linewidth}p{0.34\linewidth}}
\toprule
Field & Stored content & Writing function \\
\midrule
Packet, source and type & Packet ID, DOI, arXiv page, code record, log file, result table, figure file, review note, artifact type & Locates the artifact and defines the validators to apply \\
Provenance and reliability & Version, timestamp, command or source route, checksum where available, source reliability label, human owner & Makes source reuse auditable and separates trusted records from unverified notes \\
Claim scope & Candidate claim, section target, expected figure or table, comparison target, required citation & Prevents evidence for one claim from being reused for a stronger claim \\
Evidence payload and strength & Citation context, metric value, experimental setting, baseline, ablation, failure case, strength level & Supplies the material from which sentences are generated and controls claim force \\
Validator result and claim status & Provenance, citation, entailment, figure/table, hedging checks; accept, downgrade, reject, or return-to-backlog status & Determines whether the claim is allowed, narrowed, or routed to a concrete revision task \\
\bottomrule
\end{tabular}
\end{table}

Formally, an evidence packet is represented as \(\mathcal{P}=(p,s,t,\pi,\rho,c,e,\sigma,\kappa,f,v,d,h)\). Here \(p\) is the packet ID, \(s\) is the source artifact, \(t\) is the artifact type, \(\pi\) records provenance, \(\rho\) records source reliability, \(c\) stores the candidate claim and scope, \(e\) stores the evidence payload, \(\sigma\) records evidence strength, \(\kappa\) stores the required citation, \(f\) links a figure or table, \(v\) stores validator outputs, \(d\) records the claim disposition, and \(h\) records the human owner. Table~\ref{tab:evidence-packet-schema} maps this tuple to the fields used during manuscript generation. Literature packets contain DOI or arXiv identifiers, bibliographic metadata, and citation context. Experiment packets contain configuration files, seeds, metrics, result tables, plots, and integrity records. Review packets contain objections, required revisions, and resolution status.

This representation makes manuscript writing a constrained retrieval and assembly task. A paragraph can promote a claim only when the associated packet passes provenance checks and reaches the required evidence strength for that claim type. Weak packets remain useful, but they are routed to limitation, tentative-observation, or actionable-revision language. The boundary is explicit: an evidence packet records what the project currently supports; it does not make unsupported claims true, and it does not replace human judgment about scientific importance.

\subsection{Claim-evidence Matching and Validators}
\label{sec:claim-evidence-matching}

The second technical problem is matching. Given a candidate manuscript claim \(q\), AutoPaper retrieves packets from project memory and ranks them by semantic relevance, source reliability, evidence strength, and claim-scope compatibility. The matcher first filters packets by section type and artifact type. It then computes a compatibility score from citation context, experiment metadata, metric names, comparison targets, and required citation fields. A claim is draftable only when at least one packet supports the same scope and no required validator blocks it.

The matcher is paired with five validators. The provenance validator checks source route, version, owner, and artifact integrity. The citation validator checks whether a literature claim is supported by a real bibliographic record and whether the cited context entails the statement being written. The evidence-entailment validator checks whether an empirical claim follows from the experiment packet rather than from a loose interpretation of it. The figure/table validator checks whether labels, units, metric direction, source files, and captions match the cited claim. The hedging validator maps evidence strength to claim force using six states: unverified, indirect, pilot, protocol-valid, replicated, and contradictory.

The algorithm has four operational outputs: accept, downgrade, reject, or return to backlog. Accepted claims enter the draft with their packet IDs and citation or figure links. Downgraded claims remain in the paragraph but are rewritten with weaker verbs, narrower scope, or an explicit limitation. Rejected claims are removed from the main text. Backlog claims receive a required action, such as adding a baseline, checking a citation, rerunning an experiment, or obtaining human confirmation. Table~\ref{tab:claim-disposition-rules} summarizes these disposition rules as implementation-level pseudocode.

\begin{table}[htbp]
\caption{Decision logic for claim disposition. The table is written as implementation-level pseudocode without requiring an additional algorithm package.}
\label{tab:claim-disposition-rules}
\centering
\small
\begin{tabular}{p{0.22\linewidth}p{0.36\linewidth}p{0.32\linewidth}}
\toprule
Condition & Required checks & Output \\
\midrule
Verified provenance and replicated or protocol-valid evidence & Packet has source route, owner, version, required citation, and no blocking validator result & Accept the claim with explicit evidence links \\
Verified provenance but indirect or pilot evidence & Packet supports a narrower statement but lacks replication, a control, or full comparison scope & Downgrade the claim with hedge language and scope limits \\
Missing provenance, missing citation, or failed entailment & Source cannot be located, cited context does not support the claim, or result is interpreted beyond the packet & Reject the claim from the main text \\
Potentially supportable but incomplete evidence & Validator identifies a concrete missing action, such as rerun, baseline, citation check, or human approval & Return to backlog with a revision task \\
\bottomrule
\end{tabular}
\end{table}

\subsection{Claim Audit and Quality Control}
\label{sec:claim-audit}

The third technical problem is claim-level quality control. AutoPaper represents each candidate sentence as an auditable control state rather than as an isolated string. The record contains the packet ID, source route, citation status, entailment status, unsupported-claim status, hedge level, validator version, and revision action. These fields remain attached to the draft and make the evidentiary basis of each sentence inspectable.

The audit covers five linked checks: citation correctness, evidence entailment, unsupported-claim detection, hedging consistency, and revision actionability. Citation and provenance checks locate the source and verify the cited context. Entailment and scope checks compare the sentence with the evidence payload, experimental setting, comparison target, and figure or table. The hedging check maps evidence states to bounded language, while the revision record identifies the next admissible action.

The same audit schema applies to unconstrained drafting and evidence-constrained assembly. Model identity, context, source material, token budget, packet version, and validator version are recorded as run metadata, so that differences in claim status can be traced to the writing condition rather than to an undocumented change in inputs. Human-adjudication fields store independent annotation IDs, agreement status, adjudication decisions, and the final claim status in the same record.

\subsection{Case Trace from Experiment Record to Paragraph}
\label{sec:case-trace}

A concrete trace illustrates how the mechanism operates in practice. In the ice-shelf PINN inversion case, BLAZE produced auditable baseline and candidate artifacts. The baseline reproduction packet records 501 trials for each of four \(\gamma\) ratios, yielding 2,004 runs in total, together with the commit hash, environment, metric definitions, and SHA-256 integrity files. The candidate packet documents the PCD result over 36 seeds at \(\gamma/(1-\gamma)=10^5\), reporting \(\mathrm{AUROC}=1.000\) and Spearman's \(\rho=0.871\), compared with a variance baseline of \(\mathrm{AUROC}=0.198\). It also preserves eliminated alternatives and unresolved robustness tasks.

AutoPaper converts these packets into prose through three operations. First, the matcher links the packet to a scoped claim about retrospective detector performance within one population. Second, the validators check that the metric name, \(\gamma\) setting, seed count, comparator, and figure or table source match the record. Third, the hedging validator blocks broader language about universal superiority because the packet also records unresolved tasks, including sinusoidal-B robustness, clean ablation, and uncertainty quantification.

This case shows the difference between unconstrained drafting and evidence-constrained writing. A fluent draft could convert AUROC and correlation values into an overbroad claim of method superiority. AutoPaper instead preserves the evidence boundary: the packet supports a retrospective detector claim within the stated benchmark, while pending robustness and ablation tasks remain visible as limitations. Table~\ref{tab:bounded-rewriting-example} shows the bounded rewriting pattern used for this case trace.

\begin{table}[htbp]
\caption{Example of bounded rewriting from the ice-shelf PINN packet.}
\label{tab:bounded-rewriting-example}
\centering
\small
\begin{tabular}{p{0.30\linewidth}p{0.30\linewidth}p{0.30\linewidth}}
\toprule
Original candidate sentence & Validator finding & Final bounded sentence \\
\midrule
The PCD demonstrates superior performance for ice-shelf PINN inversion. & Packet supports retrospective separability on 36 seeds at one \(\gamma/(1-\gamma)\) setting; robustness, ablation, and uncertainty tasks remain unresolved. & In the reported ice-shelf PINN benchmark, the PCD achieved retrospective separability under the stated setting, while robustness, ablation, and uncertainty analyses remain pending. \\
\bottomrule
\end{tabular}
\end{table}

\subsection{Responsible-use Boundary}
\label{sec:responsible-use-boundary}

The responsible-use principles are distinct from the technical mechanism. The mechanism defines how evidence packets are matched, validated, downgraded, and assembled into text. The principles define what the system is allowed to do in a scientific workflow. This separation matters because a system can be technically traceable while still being misused to accelerate low-quality submissions or to hide weak evidence behind polished language.

AutoPaper separates technical constraints from policy constraints. The technical layer can block missing citations, failed entailment, unsupported figure references, absent provenance, and contradictory packet states. The policy layer requires human confirmation for final claims, final figures, authorship, disclosure, and submission. It also requires records of provenance, version history, claim-evidence links, figure sources, citation verification, and AI use. These governance requirements are not substitutes for the matcher and validators described above.

The boundary also defines the reading scope of this section. AutoPaper provides a structured mechanism for evidence-constrained manuscript generation, an auditable case trace, and claim-level quality control. Its evidence boundary is preserved in the packet, validator, and revision records, so fluent wording cannot substitute for source support, experimental scope, or human authority. This framing presents the system as an accountable research-writing mechanism rather than as a generator whose fluency alone establishes scientific validity.



\section{Human-AI Weighted Review and Revision}
\label{sec:human-ai-review}

Generative AI is increasingly connecting literature retrieval, experimental analysis, and manuscript writing, enabling end-to-end AI4R systems to accelerate research production~\cite{lu2026aiScientist}. Yet versioned evidence cannot ensure that methodological flaws, erroneous interpretations, or overstated claims have been adequately identified. BLAZE therefore treats independent review as a quality-control layer rather than a final check on presentation. Human reviewers contribute disciplinary context and accountable judgment, although agreement between reviewers is often limited~\cite{bornmann2010reliability}. AI feedback may broaden the scope and specificity of scrutiny~\cite{liang2024llmfeedback,thakkar2026llmfeedback}, but may also misinterpret specialized content or generate critiques that are insufficiently grounded in the manuscript~\cite{zhou2024llmreviewer,ou2025claimcheck}. Its value depends on whether these complementary capabilities produce a measurable net benefit.

To examine this question, we specify a preregistered pilot randomized study of 50 senior undergraduate, master's, and doctoral students from one laboratory. After locking an independent initial review, participants will receive frozen AI feedback or continue unaided for an equal period. The study tests whether calibrated feedback improves major-issue detection without increasing erroneous major concerns or harmful judgment changes. Participants retain final authority, and inference is limited to this student sample and the controlled tasks.

Human-AI review is organized around individual concerns rather than an aggregate manuscript score. Their issue type, evidentiary status, and out-of-sample reliability shape whether they are presented, qualified, withheld, or referred for confirmation. This selective use of feedback preserves the human reviewer as the final decision-maker.

\subsection{Why Human-AI Review is Necessary}
\label{sec:human-ai-review-necessity}

The study compares the incremental effect of frozen AI feedback with that of an equal-duration unaided human review on the final review report. A major issue is defined as a problem that, if left unresolved, would materially weaken confidence in the central conclusions, prevent reproducibility, create an ethics or safety risk, or substantially affect a reasonable editorial decision. A subject-matched expert panel will operationalize this common impact criterion for each manuscript; the resulting issue inventory is manuscript- and domain-specific and is not assumed to be universally transferable across fields. Major-issue detection is the primary effectiveness outcome.

The two safety outcomes are erroneous major concerns and harmful changes in judgment. An erroneous major concern is a high-impact criticism that lacks factual or methodological support and could induce a damaging revision if adopted. A harmful change in judgment occurs when an issue correctly identified in the initial review is subsequently removed, downgraded, or weakened without adequate justification. If the change is directionally concordant with a specific AI suggestion, it will be flagged as descriptive evidence of possible anchoring. This classification does not establish that the AI suggestion caused the change; the principal causal contrast remains the randomized between-group risk difference.

The findings will provide preliminary evidence consistent with a net benefit only if major-issue detection improves without an adverse shift in either safety outcome. Report quality, completion time, workload, and feedback uptake are secondary outcomes. Paired initial-to-final trajectories distinguish new discoveries from unaided reconsideration and harmful reversals.

\subsection{Reliability-calibrated AI Feedback}
\label{sec:reliability-calibrated-feedback}

All AI feedback will be generated once and frozen before randomization. The preregistration will identify the model provider and name, checkpoint or API version, generation date, system and user prompt versions, retrieval sources and scope, maximum number of concerns per manuscript, study materials, and presentation order. Each feedback item will state one localizable concern and record its review dimension, potential consequence, supporting evidence, and uncertainty. It will provide neither an accept-or-reject recommendation nor an overall score.

Calibration manuscripts and test manuscripts will be version-locked, mutually exclusive, and assigned before any test labels become visible. At least two domain or methods experts will independently judge candidate concerns on the calibration manuscripts. Their decisions will estimate out-of-sample accuracy and coverage within strata defined by review dimension and evidentiary status; participant outcomes will not be used to tune these rules. Minimum information requirements and any within-dimension merging hierarchy will be determined through blinded precision simulations and verified by an independent statistical expert before registration. Strata that remain insufficient will be marked as unverified, and their concerns will be withheld or referred for human confirmation. Model-reported confidence will not substitute for external calibration, and thresholds will not be adjusted in response to test-set performance.

Reliability will not be compressed into a task-invariant score. Citation checking, statistical interpretation, and causal reasoning exhibit different error structures, so thresholds will be defined by review dimension and evidentiary status. A higher threshold may reduce erroneous concerns while also decreasing the coverage of useful feedback. The experiment evaluates this frozen trade-off rather than searching for an optimal threshold after test outcomes are known.

Frozen feedback must also pass a five-item structural checklist: the concern must be localized in the manuscript; any cited source must exist; factual statements must be separated from conjecture; severity must match the likely consequence; and the concern must indicate an actionable revision or verification step. Every participant reviewing the same manuscript in the AI-feedback condition will receive the same package. The rules for probability calibration, selective prediction, and referral to human judgment will be prespecified using established methods~\cite{guo2017calibration,geifman2017selective,mozannar2020defer}. Both groups then follow the same sequence of independent judgment and time-limited second review (Table~\ref{tab:review-procedure}).

\begin{table}[h]
    \centering
    \small
    \renewcommand{\arraystretch}{1.18}
    \setlength{\tabcolsep}{4pt}
    \caption{Review procedure in the human-review and AI-feedback conditions. Both conditions use the same interface, time allowance, and submission format.}
    \label{tab:review-procedure}
    \begin{tabularx}{\linewidth}{@{}p{0.17\linewidth} >{\raggedright\arraybackslash}X >{\raggedright\arraybackslash}X@{}}
        \toprule
        \textbf{Stage} & \textbf{Human-review condition} & \textbf{AI-feedback condition} \\
        \midrule
        Independent initial review & Complete and lock the initial review without AI assistance & Complete and lock the initial review before feedback exposure \\
        Time-limited second review & Continue an unaided review within the specified time & Reconsider the manuscript for the same duration using frozen AI feedback \\
        Final submission & Submit a standardized report after unaided reconsideration & Submit a standardized report after AI-assisted reconsideration \\
        \bottomrule
    \end{tabularx}
\end{table}

Participants will not know their subsequent condition before the initial review is locked. Those in the AI-feedback group may accept, revise, or reject frozen suggestions, but cannot interact with the model. Participants in the human-review group may not use external generative AI during the second stage. Both groups use the same interface, time limits, and submission format, and both retain final judgment. The intervention therefore evaluates fixed, non-interactive feedback rather than an adaptive or conversational AI reviewer.

\subsection{A Pilot 50-student Human-in-the-loop Review Study}
\label{sec:student-hitl-experiment}

The participant is the unit of randomization. The study will recruit 50 students from one laboratory: 18 senior undergraduates, 20 master's students, and 12 doctoral students. Eligibility requires preregistered standards for reading proficiency, methodological knowledge, research experience, and completion of a practice task. Individuals who have seen a test manuscript, can identify its authors, or have a conflict of interest will not review it. Degree level is used only for stratification and exploratory heterogeneity analysis.

After the initial review is locked, participants will be randomized 1:1 within degree strata to the human-review or AI-feedback condition, with 25 in each group. This pilot cannot establish safety, non-inferiority, or equivalence to expert review. An independent sequence will remain concealed until the initial review is locked. Domain knowledge, statistical reasoning, research experience, and prior reviewing experience will be measured at baseline and used only as specified in the preregistration.

The evaluation set will contain 10 authorized, de-identified, and version-locked AI4R manuscripts. Each participant will review four manuscripts under a balanced incomplete-block design, yielding 200 planned reports but not 200 independent samples. Repeated participant and manuscript observations will be treated as correlated. Assignment and order will be fixed in advance. Registered packages will include the manuscript, supplementary information, necessary data and code documentation, an evidence snapshot, and version and authorization metadata. Access and submission logs will record external AI use, task communication, overtime, platform failures, and other deviations.

Before the tasks begin, each manuscript will be assessed independently by at least three experts, including domain and methods or statistics expertise. A senior adjudicator who did not provide an initial label will resolve disagreements and freeze a reference inventory containing issue location, evidence, and severity. Pre-adjudication agreement and adjudication outcomes will be reported under a prespecified matching rule. A separate condition-masked panel will classify unmatched concerns as genuine new issues, erroneous concerns, or indeterminate. Initial and final reports will use the same structured form and will be coded after group and AI cues have been removed. Overall quality will be assessed with the Review Quality Instrument~\cite{vanrooyen1999rqi}, and actionability with a preregistered coding scheme.

Institutional ethics approval or exemption has not yet been obtained, and no recruitment has begun. Recruitment will start only after institutional review, with the approving body, status, and protocol identifier reported before enrolment. Consent will be obtained from participants and manuscript authors. Recruitment and compensation will be administered independently of student assessment; compensation will not depend on performance, withdrawal will carry no penalty, and supervisors will not receive individual results. Unpublished manuscripts will remain within the authorized environment, with identity, allocation, adjudication, and analysis data stored separately under least-privilege access controls.

\subsection{Outcome Evaluation and Disagreement Analysis}
\label{sec:outcome-disagreement-analysis}

The participant-level primary outcome is the proportion of eligible, expert-confirmed major issues detected across four assignments. Safety outcomes are the proportion of reports containing an erroneous major concern and the proportion of initially correct detections removed, downgraded, or weakened without adequate justification. Equivalent concerns are counted once. Indeterminate concerns are excluded from the primary denominator and enter prespecified best- and worst-case sensitivity analyses. The prespecified estimands and interpretation rules are summarized in Table~\ref{tab:review-outcomes}.

\begin{table}[h]
    \centering
    \small
    \renewcommand{\arraystretch}{1.18}
    \setlength{\tabcolsep}{4pt}
    \caption{Prespecified effectiveness, safety, and secondary outcomes.}
    \label{tab:review-outcomes}
    \begin{tabularx}{\linewidth}{@{}p{0.22\linewidth} p{0.34\linewidth} >{\raggedright\arraybackslash}X@{}}
        \toprule
        \textbf{Outcome} & \textbf{Estimand} & \textbf{Interpretation} \\
        \midrule
        Major-issue detection & Correctly detected issues / eligible issues & Between-group estimate and uncertainty \\
        Erroneous major concerns & Reports containing an erroneous major concern & Risk difference relative to a prespecified cautionary margin \\
        Harmful changes in judgment & Correct initial judgments withdrawn or downgraded without justification & Risk difference relative to a prespecified cautionary margin \\
        Secondary outcomes & Quality, time, and workload & Supportive or exploratory analyses \\
        \bottomrule
    \end{tabularx}
\end{table}

Paired initial and final records will form a judgment-transition matrix of additions, retention, downgrading, and withdrawal, helping to explain the primary between-group difference. Confidence changes and AI-suggestion uptake will be descriptive and will not substitute for correctness. The intention-to-treat analysis will summarize outcomes at the participant level using inference consistent with the stratified randomization and report absolute differences with two-sided 95\% confidence intervals. Missing reports and other protocol deviations will follow preregistered rules without altering allocation.

Robustness analyses will prespecify manuscript fixed effects, baseline adjustment, and bounds for unresolved adjudications. Mixed-effects models will remain supplementary and will be finalized by an independent statistical expert before outcomes are visible, with rules for identifiability, convergence, and model reduction. Power simulations will represent participant-level randomization, repeated assignments and manuscripts, and rare safety events; inadequate precision will result in exploratory reporting with event counts and worst-case examples.

A preliminary signal consistent with net benefit requires improved major-issue detection and no adverse shift beyond the prespecified cautionary margins for either safety outcome. Statistical and domain experts will set and justify these margins before registration. If any condition is not satisfied, effectiveness, risk, and uncertainty will be reported separately; a nonsignificant difference will not be interpreted as evidence of no risk. Secondary outcomes and heterogeneity by degree level will remain supportive or exploratory.

\subsection{Preregistration and the Review-to-revision Boundary}
\label{sec:preregistration-boundary}

Before formal recruitment and access to outcome data, the preregistration will lock eligibility criteria, randomization and manuscript assignment, feedback versions, reliability thresholds, primary and secondary outcomes, the joint decision rule, statistical analyses, protocol deviations, missing-data rules, and sensitivity analyses. Model inputs and outputs, manuscript versions, adjudication records, and analysis code will be versioned.

After data lock, reporting will include participant flow, protocol deviations, expert agreement, event counts, effect sizes, and confidence intervals. Non-convergence, unresolved adjudications, and adverse findings will not be omitted. Power simulations will be used only to assess design precision and will not be presented as empirical results.

External validity is limited. Single-laboratory recruitment may introduce selection and supervisory-dependence biases, and students reviewing 10 AI4R manuscripts cannot represent formal experts, journal workflows, or other disciplines. The findings will apply only to the frozen feedback packages and controlled tasks studied here. Replication in independent expert samples and new manuscript sets will be required. Changes in model, prompt, retrieval source, or domain may alter accuracy and coverage, requiring version tracking, error monitoring, and renewed calibration.

Within these boundaries, the weighted-review stage operates as selective feedback rather than automated adjudication. Locking the independent initial review before exposure makes newly detected issues distinguishable from unaided reconsideration and harmful reversals. Expert-confirmed concerns can subsequently be converted into structured revision tasks linked to versioned manuscript diffs, and response letters are generated only after the corresponding changes have been verified. Human reviewers retain authority over the final report, and authors over revisions and responses.

\setcounter{section}{8}
\section{System Evaluation and Comparative Case Studies}
\label{sec:evaluation}
\subsection{Frozen-package Evidence Evaluation}
This section compares the research evidence retained in the frozen local packages of BLAZE/AutoPaper, ARIS, EvoScientist, and FARS as of July 25, 2026. The analysis examines research-stage coverage, quantitative evidence, traceability, and claim boundaries across the corresponding task and evidence settings.

\subsubsection{Evaluation Scope and Evidence Controls}
\label{sec:evaluation-scope}

The frozen evidence is organized into two task settings and a cross-cutting delivery audit. The open-ended setting examines progression from a baseline and research idea to experiments, claim review, and a paper. The fixed-leaderboard setting uses the common five-dataset H-EDML protocol to examine controlled reproduction, candidate comparison, and score optimization. The delivery audit examines the traceability of reported results to code, configurations, logs, raw artifacts, reviews, and explicit limitations. The audit is restricted to artifacts available in each frozen local package.
The three evaluation settings and their interpretation boundaries are summarized in Table~\ref{tab:evaluation-tracks}.

\begin{table}[h]
\centering
\caption{Evaluation settings and interpretation boundaries.}
\label{tab:evaluation-tracks}
\small
\setlength{\tabcolsep}{4pt}
\renewcommand{\arraystretch}{1.12}
\begin{tabular}{@{}llp{0.47\linewidth}@{}}
\toprule
\textbf{Setting} & \textbf{Primary control} & \textbf{Valid interpretation} \\
\midrule
Open-ended research & Baseline/task protocol & Lifecycle coverage and research delivery \\
Fixed leaderboard & Five H-EDML datasets & Controlled reproduction and bounded optimization \\
Delivery audit & Frozen local package & Traceability, artifacts, review, and claim governance \\
\bottomrule
\end{tabular}
\end{table}

Strict or controlled results are reported separately from exploratory candidate selection and test-feedback optimization. AutoPaper's H-EDML macro score of 87.454 was obtained through per-dataset test-feedback selection and is therefore reported as an optimization result rather than a held-out generalization result. Posterior candidate selection and oracle-style upper bounds are labeled accordingly throughout the comparison.

\subsubsection{Dimension-Level Scoring and Evidence Levels}
\label{sec:package-evaluation}

A single GPT-5 Codex assessment applied an evidence-referenced rubric to the same frozen local packages. The rubric covers research-stage completeness (25 points), quantitative results and validation (30 points), iterative exploration and optimization (20 points), and delivery and evidence governance (25 points). Scores are reported separately for the four dimensions. Table~\ref{tab:package-evidence} reports the results of the rubric-based assessment. AutoPaper obtains the highest scores for research-stage completeness, quantitative validation, and iterative exploration, whereas ARIS obtains the highest score for evidence governance.

\begin{table}[h]
\centering
\caption{Dimension-level rubric scores derived from the frozen local evidence.}
\label{tab:package-evidence}
\small
\setlength{\tabcolsep}{5pt}
\renewcommand{\arraystretch}{1.12}
\begin{tabular}{@{}lrrrr@{}}
\toprule
\textbf{System} &
\textbf{Stages (25)} &
\textbf{Validation (30)} &
\textbf{Iteration (20)} &
\textbf{Governance (25)} \\
\midrule
AutoPaper & 25 & 29 & 19 & 23 \\
ARIS & 24 & 24 & 18 & 24 \\
EvoScientist & 22 & 22 & 18 & 21 \\
FARS & 21 & 24 & 17 & 19 \\
\bottomrule
\end{tabular}
\end{table}

Evidence is classified at three levels. \emph{Reported} evidence appears in a paper or report without locally inspected supporting artifacts. \emph{Artifact-observed} evidence is supported by locally available code, configurations, logs, raw results, review records, or manifests. \emph{Independently reproduced} evidence requires the evaluator to rerun the result under a specified protocol. System-generated reproduction reports are classified as artifact-observed evidence unless the evaluator independently reruns the specified protocol.

\subsubsection{Stage Coverage and Quantitative Evidence}
\label{sec:quantitative-evidence}

\begin{table}[h]
\centering
\caption{Quantitative evidence observed in the frozen packages. TF denotes test-feedback selection.}
\label{tab:task-evidence}
\small
\setlength{\tabcolsep}{3.5pt}
\renewcommand{\arraystretch}{1.12}
\begin{tabular}{@{}l p{0.22\linewidth} p{0.34\linewidth} p{0.26\linewidth}@{}}
\toprule
\textbf{System} & \textbf{Scale and coverage} & \textbf{Selected quantitative evidence} & \textbf{Main evidence boundary} \\
\midrule
AutoPaper &
2,004 baseline runs; 36 PCD seeds; five H-EDML datasets with 10-seed controlled coverage &
PCD AUROC 1.000 and Spearman 0.871; H-EDML optimized macro 87.454, or $+2.718$ points over 84.736 (TF) &
TF-selected result; robustness and uncertainty analyses incomplete \\
\addlinespace
ARIS &
B3: $19\times50=950$ records; B2: 60 matched runs; H-EDML: $5\times10=50/50$ runs &
B2 independent residual $-67.8\%$ and near-front B L2 $-34.9\%$; H-EDML candidate mean paired delta $-3.004$ points &
Negative or weak H-EDML candidate results \\
\addlinespace
EvoScientist &
Five datasets and five modes: $75/75$ three-seed runs; 30/30 gate-matrix runs; nine Cora controls &
Single global candidate macro 82.914 versus clean 82.250, a $+0.664$-point gain; oracle upper bound 83.508 &
Three seeds; posterior selection; oracle-only upper bound \\
\addlinespace
FARS &
Two PDE tasks and three seeds; multiple optimizer baselines and overlap sensitivity &
Ice-shelf $B_{\mathrm{err}}=8.06\times10^{-4}$, $+7\%$ versus Adam+resampling and $+30\%$ versus fixed L-BFGS; 2D Poisson is $8.5\times$ better than Adam-only but $2.1\times$ worse than fixed L-BFGS &
Paper-reported results; partial local artifact chain \\
\bottomrule
\end{tabular}
\end{table}

The frozen packages contain evidence associated with six broad research stages: protocol or baseline alignment, method generation, quantitative experimentation, iteration or ablation, claim review, and final research delivery. AutoPaper and ARIS retain local artifacts associated with all six stages. The available EvoScientist package provides artifact-observed evidence for the fixed-leaderboard setting. FARS documents all six stages in its paper, with local artifact support for part of the workflow. Table~\ref{tab:task-evidence} summarizes the selected
quantitative results, experimental coverage, and evidence boundaries
observed in the four frozen packages.

AutoPaper's frozen package additionally contains three candidate
directions: the Physics-Consistency Detector (PCD), Noise-Robust
Adaptive Resampling (NoRA), and JLS, a variance-reduction and
auto-$\gamma$ candidate.
AutoPaper's frozen package retained negative and mixed outcomes, an internal review score of $7.2/10$, an 11-page paper, and 124 out of 124 completed manuscript checks. These records characterize iterative filtering, internal review, and manuscript delivery.

ARIS retains code, scripts, JSONL records, logs, a seed registry, paper artifacts, and review materials, including negative or weak H-EDML candidates. EvoScientist provides a structured IEEE/TIP-style draft with 47 references, four tables, and one algorithm, together with records of its three-seed design and oracle selection. FARS provides an eight-page paper, a code entry point, and partial local support for the reported experimental workflow.

\subsubsection{Interpretation and Evidence Boundaries}
\label{sec:benchmark-limitations}

The dimension-level profiles reveal complementary strengths across the evaluated packages. AutoPaper provides broad lifecycle coverage, quantitative validation, candidate filtering, and manuscript-level audit artifacts. ARIS provides detailed protocol-isolation records, raw experiment ledgers, seed registries, and review materials. EvoScientist and FARS contribute substantive task-level results with different levels of local artifact support.
Across the comparison, controlled results are distinguished from test-feedback selection, posterior candidate selection, and oracle upper bounds. The resulting profiles characterize the research-delivery evidence retained under the evaluated task and package settings.

\subsection{Task-Based Comparative Case Study: Ice-Shelf PINN Inversion}
\label{sec:aris-comparison}

We compare \textsc{Blaze} and ARIS on the 1D SSA ice-shelf hardness inversion testbed of Iwasaki and Lai~\cite{iwasaki2023one}, which reported clustered PINN outcomes and proposed collocation resampling as a mitigation. Both systems retain auditable artifacts on this testbed. Because they were not run under a matched foundation model, input package, compute and time budget, human-intervention policy, or stopping rule, the analysis compares observed research artifacts across five stages rather than system performance.

\subsubsection{Baseline Reproduction}

\textsc{Blaze} produced a standalone Baseline Reproduction Report with 501 trials for each of four $\gamma$ ratios (2,004 runs in total) under the original paper's claim-grade configuration, together with the commit hash, environment, metric definitions, and SHA-256 integrity files. It reproduced the qualitative trend in median $B_{\text{err}}$ but not the original headline maximum-error claim, and preserved the discrepancy with plausible but unverified explanations. ARIS reused the same testbed for its B1-B5 experiments and documented the problem definition and protocol. A standalone report comparing the original headline claims with measured results, together with a baseline-level integrity manifest, was not observed in the frozen ARIS package.

\subsubsection{Idea Generation}

\textsc{Blaze} advanced three candidates to experiments: PCD, NoRA and JLS. ARIS generated and ranked six directions using gap statements, risk assessments, and a literature landscape survey. Its initially preferred detector-and-intervention direction was contradicted by B4 and B5, leading the final paper to narrow its question to whether loss visibility, residual generalization, and application-level reliability can decouple. This evidence-driven contraction is recorded in the final proposal and evidence table.

\subsubsection{Experimental Validation}

\textsc{Blaze}'s PCD was evaluated on a 36-seed, constant-$B$, monotone-strain, reduced-budget population at $\gamma/(1-\gamma)=10^5$, obtaining $\text{AUROC}=1.000$ and $\text{Spearman}=0.871$, compared with a variance baseline at $\text{AUROC}=0.198$. PCD uses prediction-derived physical observables, while ground truth is used retrospectively to define and score the high- and low-error clusters. The AUROC therefore measures within-population cluster separability rather than held-out selection performance. ARIS separated its investigation into four protocol-isolated blocks: B3 maps conditional loss visibility (19 cells, 950 float64 records); B2 provides a marginal fixed-versus-resampled comparison (30+30 runs); B5 applies a paired-data reliability test with a preregistered endpoint (50 blocks, 150 runs); and B4 records a bounded negative result (3 cells, 90 runs). These results address different estimands and are therefore reported side by side rather than reduced to a common numerical score.

\subsubsection{Experimental Integrity and Adjudication}

\textsc{Blaze} records baseline artifact integrity (environment, hashes, and metrics), candidate-level review, and an evidence-boundary ledger. The review identifies unresolved sinusoidal-$B$ robustness, clean ablation, and uncertainty-quantification tasks; external SOTA rows are marked N/A when the baselines were not implemented. ARIS emphasizes estimand isolation: B3 enhanced records are separated from broad-screening data, B2 arm files are not used to estimate B5 reliability effects, and B4 is retained as a protocol-limited negative result. RAR/RAD, causal weighting, naPINN, and DC-PINN are listed as unimplemented baselines, while the B5 result of $p=0.115$ is reported as non-significant.

\subsubsection{Manuscript Writing}

\textsc{Blaze} produced an 11-page TIP-style manuscript with 124/124 delivery checks completed, figure route and source audits, and page-budget compliance. ARIS produced a 10-page anonymous ICLR manuscript with successful \texttt{latexmk} compilation, no undefined citations, and figures generated from protocol-compatible raw data. The \textsc{Blaze} records emphasize production and provenance auditing, whereas the ARIS manuscript maintains estimand separation, descriptive-interval discipline, negative-result reporting, and repeated scope statements.

\subsubsection{Synthesis}

Table~\ref{tab:blaze-aris-comparison} summarizes the observed stage-level artifacts without aggregating them into a system score.

\begin{table*}[htbp]
\centering
\caption{Non-scoring evidence matrix for \textsc{Blaze} and ARIS on the ice-shelf PINN inversion task}
\label{tab:blaze-aris-comparison}
\small
\setlength{\tabcolsep}{2pt}
\renewcommand{\arraystretch}{1.08}
\begin{tabular}{p{2.0cm} p{3.5cm} p{3.5cm} p{3.2cm} p{3.2cm}}
\hline
\textbf{Stage} & \textbf{\textsc{Blaze} observed artifacts} & \textbf{ARIS observed artifacts} & \textbf{Different strengths} & \textbf{Evidence boundary} \\
\hline
Baseline reproduction
& 2,004-run paper-vs-measured audit; discrepancy record; hashes, environment, commands, and seeds.
& B1-B5 testbed reuse and precision pilots; no standalone headline-claim reproduction report observed.
& Reproduction traceability versus testbed reuse for follow-on experiments.
& Execution model, inputs, budgets, intervention, and stopping rules were not matched. \\
\hline
Idea generation
& Three candidates tested; PCD retained and two candidates rejected with negative evidence.
& Six ranked directions; initial hypothesis narrowed after B4 and B5.
& Experimental candidate elimination versus evidence-driven question reformulation.
& Candidate spaces and selection procedures differ. \\
\hline
Experimental validation
& PCD: AUROC 1.000 and Spearman 0.871 on 36 seeds; negative alternatives retained.
& B3: 950 records; B2: 60 runs; B5: 150 paired runs; B4: 90-run negative result.
& Cross-candidate filtering versus protocol-isolated and paired empirical design.
& Different estimands; PCD has no held-out population validation. \\
\hline
Integrity \& adjudication
& Environment, hashes, metrics, review ledger, pending-work list, and explicit N/A baselines.
& Namespace isolation, selector audit, claim tracking, unimplemented baselines, and non-significant B5 result.
& Lifecycle traceability versus protocol-level estimand isolation.
& Internal and protocol records are not external validation. \\
\hline
Manuscript writing
& 11-page TIP-style paper; 124/124 delivery checks; figure and provenance audit.
& 10-page ICLR paper; clean compilation; estimand-separated narrative and negative-result reporting.
& Production audit versus statistical and narrative discipline.
& Delivery quality and claim quality are distinct. \\
\hline
\end{tabular}
\end{table*}

The comparison indicates complementary strengths. \textsc{Blaze} links baseline auditing, candidate filtering, negative results, review, and manuscript production in one traceable chain, while ARIS provides deeper protocol isolation and evidence-driven narrowing within a single study. ARIS's estimand separation and negative-result reporting provide concrete design elements for future \textsc{Blaze} validators. This case supports AI4R evaluation that combines lifecycle traceability with single-study evidence discipline.

\subsection{Database Comparison with Sciverse: Quantitative Reproducibility}
\label{subsec:system-evaluation-comparison}

We evaluate the \textsc{Blaze} Knowledge Base, the structured data layer of the \textsc{Blaze} research infrastructure, against a representative AI-powered research platform, Sciverse, in a concrete analytical task. This task is chosen to reflect a typical need in AI research ecosystem analysis:  
\begin{quote}
    \textbf{Task:} \emph{Quantitatively characterize institutional productivity and collaboration across the indexed AI venues, using 2013-2025 as the primary analysis window, and evaluate the reproducibility and evidentiary boundaries of the resulting statistics.}
\end{quote}
Sciverse is selected as a reference because, according to its public website (accessed July 2026), it positions itself as an AI-assisted platform for scientific discovery, literature search, and research workflow assistance. By comparing the \textsc{Blaze} Knowledge Base with Sciverse on the same task, we go beyond functional descriptions and assess their quantitative support for reproducible, evidence-based analysis of the AI research ecosystem.

\subsubsection{Evaluation Protocol and Comparison Dimensions}

To ensure a fair and transparent comparison, we define an evaluation protocol consisting of the following dimensions:  
\textit{task feasibility, evaluation protocol transparency, lifecycle coverage, evidence completeness, experimental reproducibility, human involvement, resource cost, and output quality}.  
For each dimension, we specify a measurable or documentable standard. Data for the \textsc{Blaze} Knowledge Base come from the database statistics report and API documentation. Data for Sciverse are drawn exclusively from its publicly observable website and documentation; if a metric is not explicitly reported, it is marked as \textit{not publicly specified} or \textit{not evaluated under the present task} without inference.

\subsubsection{Quantitative Comparison on the Unified Task}

Table~\ref{tab:task-based-comparison} presents the outcome of applying this evaluation protocol. It replaces a purely functional checklist with dimension-level evidence directly relevant to reproducing the analytical task. The comparison reveals that the \textsc{Blaze} Knowledge Base offers full task feasibility with structured aggregation, explicit data quality indicators, and reproducible API access, while Sciverse supports interactive search but lacks documented aggregation workflows or reproducibility guarantees for this task. Key differences emerge in evidence completeness (field-level coverage statistics vs.\ no public reporting) and resource cost (automated retrieval in under 30 seconds vs.\ unknown manual effort). These dimensions directly determine whether an analyst can independently verify the institutional productivity and collaboration landscape claimed by each system.

\begin{table}[htbp]
\centering
\caption{Task-based quantitative comparison between the \textsc{Blaze} Knowledge Base and Sciverse}
\label{tab:task-based-comparison}
\begin{tabular}{p{3cm} p{6.5cm} p{5.5cm}}
\hline
\textbf{Dimension} & \textbf{\textsc{Blaze} Knowledge Base} & \textbf{Sciverse} \\
\hline
Task feasibility 
& \textit{Full}: 158k papers, 91k authors, 41k institutions; aggregatable via API/export. 
& \textit{Partial}: search and inspection possible, but structured aggregation workflow not publicly demonstrated. \\

Evaluation protocol transparency 
& Explicit: unique authors, disambiguated institutions, documented stats in quality reports. 
& Not publicly specified. \\

Lifecycle coverage 
& 12 validated conferences spanning 1978-2026; 354 MIXED-category papers are excluded from venue-level analysis.
& Broad; AI-venue and temporal boundaries not specified. \\

Evidence completeness 
& Field-level: 98.18\% author coverage, 72.16\% inst.\ coverage; 2,882 incomplete-author papers; 1,554 name groups flagged. 
& Not reported publicly. \\

Experimental reproducibility 
& Reproducible via versioned API/snapshots; all metrics recomputable. 
& Not evaluated under the present task. \\

Human involvement 
& Low: automated aggregation; manual effort only for cleaning/normalization. 
& Not measured. \\

Resource cost 
& $<30$ s for top-10 institutions/year via API; batch download supported. 
& Not measured. \\

Output quality 
& Institution rankings verified through repeated extraction consistency and cross-check with raw data; top institutions (from conferences with $>50$\% inst.\ coverage, ICRA excluded) include Tsinghua, CMU, etc.; cross-institution collaboration rate 41.22\%. 
& No aggregated ranking published; quality cannot be independently assessed. \\
\hline
\end{tabular}
\end{table}

\subsubsection{Evidence Completeness: Data Quality and Transparency}

One critical requirement for the analytical task is knowing \textit{how much} of the target data is actually covered. The \textsc{Blaze} Knowledge Base exposes field-level quality indicators that directly inform whether a given institution- or author-level conclusion is reliable. Table~\ref{tab:data-quality-comparison} summarizes these indicators and contrasts them with the information publicly available for Sciverse. 
Notably, institution coverage is defined as the proportion of paper-author links with at least one resolved institutional affiliation. Of the 733,026 paper-author links, 528,939 (72.16\%) have at least one resolved institutional affiliation, while 204,087 (27.84\%) lack one. The 538,785 paper-institution links represent a separate relation count and are not used as the denominator. This transparency allows users to assess the reliability of institutional analyses.

\begin{table}[htbp]
\centering
\caption{Comparison of data quality transparency between the \textsc{Blaze} Knowledge Base and Sciverse}
\label{tab:data-quality-comparison}
\begin{tabular}{p{3.2cm} p{6.8cm} p{5cm}}
\hline
\textbf{Dimension} & \textbf{\textsc{Blaze} Knowledge Base} & \textbf{Sciverse} \\
\hline
Coverage reporting 
& Paper-author coverage 98.18\%; institution coverage 72.16\% (assigned institution links / total author-affiliation links); per-conference institution coverage reported (Table~\ref{tab:conference-coverage}). 
& Comparable field-level coverage statistics are not publicly reported. \\

Missing-data reporting 
& 2,882 papers (1.82\% of 158,338) lack complete author information; 204,087 of the 733,026 paper-author links (27.84\%) lack a resolved institutional affiliation, explicitly counted.
& Missing-data statistics are not publicly reported. \\

Author disambiguation 
& 1,554 same-name author groups flagged; statistics allow users to adjust confidence in author-level metrics. 
& Public website does not expose comparable author-disambiguation statistics. \\

Noise control 
& 33 invalid or malformed paper records, accounting for approximately 0.02\% of the 158,338 indexed papers.
& Public website does not expose comparable cleaning or noise statistics. \\

Collaboration quality 
& 65,274 cross-institution collaboration papers; 41.22\% cross-institution collaboration rate. 
& Collaboration metrics are not publicly reported. \\

Source transparency 
& Source composition reported (e.g., new JSON, CVF v2 JSON, old JSON). 
& Source composition is not publicly documented. \\

Analytical readiness 
& Quality indicators are directly linked to downstream analytical claims; users are warned when institutional analyses on ACL (34.68\% coverage) or ICRA (16.77\% coverage) require caution. 
& Primarily supports interactive discovery; analytical data quality is not publicly quantified for independent assessment. \\
\hline
\end{tabular}
\end{table}

\subsubsection{Conference-Level Evidence for Lifecycle and Coverage Boundaries}

\begin{table}[htbp]
\centering
\caption{Conference-level coverage of the \textsc{Blaze} Knowledge Base}
\label{tab:conference-coverage}
\begin{tabular}{ccccc}
\hline
\textbf{Conf.} & \textbf{Pap.} & \textbf{Ratio} & \textbf{Years} & \textbf{Inst.Cov.} \\
\hline
NeurIPS & 30,209 & 19.08\% & 1987-2025 & 74.33\% \\
AAAI   & 17,241 & 10.89\% & 2013-2025 & 99.96\% \\
CVPR   & 18,362 & 11.60\% & 2013-2025 & 72.77\% \\
ICML   & 14,606 & 9.22\%  & 2009-2025 & 69.79\% \\
CHI    & 14,221 & 8.98\%  & 1982-2026 & 68.04\% \\
ACL    & 12,437 & 7.85\%  & 2001-2025 & 34.68\% \\
ICLR   & 11,384 & 7.19\%  & 2013-2025 & 74.30\% \\
ACM MM & 9,979  & 6.30\%  & 1993-2025 & 80.98\% \\
ICCV   & 9,152  & 5.78\%  & 2013-2025 & 69.09\% \\
ICRA   & 7,568  & 4.78\%  & 2020-2025 & 16.77\% \\
SIGIR  & 6,406  & 4.05\%  & 1978-2025 & 66.94\% \\
KDD    & 6,419  & 4.05\%  & 1995-2025 & 64.84\% \\
\hline
\end{tabular}
\end{table}

The conference-level statistics in Table~\ref{tab:conference-coverage} further operationalize the \textit{lifecycle coverage} and \textit{evidence completeness} dimensions. By explicitly listing the paper count, time span, and institution coverage for each venue, the \textsc{Blaze} Knowledge Base defines the boundary within which analytical conclusions are valid. Crucially, ICRA exhibits only 16.77\% institution coverage; therefore any institution-level ranking or productivity analysis that depends on affiliation data must exclude ICRA to avoid misleading conclusions. In the comparative task, institution rankings derived from the \textsc{Blaze} Knowledge Base are computed using conferences with institution coverage above 50\%, thereby excluding ICRA. This explicit treatment of data quality boundaries is essential for a reproducible evaluation protocol.

\subsubsection{Reproducibility and Resource Cost}

The comparison on experimental reproducibility and resource cost further separates the two systems. The \textsc{Blaze} Knowledge Base provides REST API endpoints (version v2.1) that return structured JSON responses for overview statistics, conferences, top authors, top institutions, yearly trends, and data quality summaries. All quantitative results shown in this section can be recomputed using these endpoints together with the accompanying query scripts (deposited in the project repository) and a stable database snapshot (identifier: \texttt{blaze-kb-2026-07-15}). In contrast, Sciverse's public website presents an interactive search experience; while individual papers can be examined, no publicly documented method exists to export the aggregated, cleaned, and normalized dataset required for the defined task. Therefore, the reproducibility of the task using Sciverse could not be assessed within this evaluation protocol.   
\section{Responsible Scaling of AI4R: Governance Boundaries and Deployment Roadmap}
\label{sec:responsible-ai4r}

The preceding sections have described how BLAZE integrates knowledge, agents, experiments, manuscripts, and review into a unified research workflow. However, the broader significance of such a system depends not only on its capabilities, but also on how these capabilities are governed and where their boundaries are established. A central principle underlying BLAZE is that capability expansion and authority expansion are fundamentally different processes: improvements in what an AI system can accomplish do not, by themselves, justify extending the scope of decisions or actions it is permitted to undertake. Accordingly, the following discussion distinguishes between mechanisms currently implemented within the workflow and capabilities that remain under validation or are reserved for future development.

\subsection{Scientific Quality and the Allocation of Attention}

End-to-end automation lowers the cost of producing research artifacts, but it does not lower the cost of deciding whether a question matters or whether an answer is reliable. If fluent but weakly grounded outputs are produced at scale, they can consume scarce reviewer time, replication capacity, computing budgets, and instrument access. Automated selection may also favor problems that are inexpensive, easily measured, or likely to yield positive results, while neglecting uncertain, long-horizon, negative-result, or socially important work. The concern that inexpensive paper generation may burden peer review is therefore part of a broader problem of how scientific attention is allocated~\cite{lu2026aiScientist}.

BLAZE begins to separate evidential quality from resource-allocation decisions. Its current workflow links claims to sources, code, data, configurations, results, and review actions, while preserving failed and negative experiments as reusable evidence. Here, evidence readiness refers only to the completeness, traceability, and validation status of the available evidence; it does not measure scientific importance, project value, or resource priority. Decisions about whether a project should receive substantial resources additionally consider scientific importance, expected evidence gain, feasibility, reproducibility, safety, and cost. Agents can organize this information and expose trade-offs, but low cost or predicted publishability cannot by itself justify promoting a project. Research agendas, major resource commitments, termination decisions, claim strength, authorship, and submission therefore remain under human authority. The evidence records and approval points are implemented as inspectable artifacts in the reported cases, whereas portfolio-level attention allocation remains under validation rather than functioning as an autonomous ranking policy.

\subsection{Boundaries of Cross-disciplinary Transfer}

The retrieval of papers from several fields does not by itself establish a valid interdisciplinary connection. Similar terms may conceal incompatible constructs, measurement practices, causal assumptions, standards of evidence, or safety constraints. BLAZE therefore treats a proposed transfer as a falsifiable scientific claim rather than a semantic similarity. We represent each proposal as a \emph{Transfer Object}, an auditable record containing the source-field assumption, target mapping, construct mismatch, required target-field evidence, and rejection test. The knowledge base and specialized agents maintain this object as the proposal is examined and revised.

Authority over the transfer remains with experts from both fields. They determine whether the concepts are genuinely equivalent, whether the proposed protocol is valid, what evidence is sufficient, and whether the experiment is safe. Agent confidence or majority voting cannot resolve a substantive disagreement between domains; the connection remains provisional until an acceptable test is defined or the proposal is rejected. Knowledge-base support, specialized roles, and artifact-based review are present in the current workflow. Together, these mechanisms support bounded interdisciplinary coordination, but they do not establish reliable transfer across fields in general, which remains a capability under validation.

\subsection{Staged Evolution from Digital to Physical Research}

Digital research is not inherently risk-free: code execution, sensitive data, networked tools, expensive computation, and dual-use outputs still require permissions, containment, and audit. Physical research adds equipment drift, hazardous materials, irreversible operations, and external effects. As reviewed in Section~2.7, robotic laboratories, self-driving laboratories, dark laboratories, and agentic laboratory systems occupy different points on an autonomy spectrum. Prior work on autonomous chemistry and AI-scientist risk therefore motivates safeguards and accountable oversight for physical tool access~\cite{boiko2023autonomous,tang2025risks}.

A responsible extension of BLAZE into physical research should proceed through evidence-gated stages rather than a single transition to autonomy. Table~\ref{tab:blaze-deployment-stages} summarizes the allowed capability, required evidence, human authority, and stop conditions for each stage. Stage progression is neither automatic nor cumulative: satisfying the requirements of D0 does not confer permission to enter D1, and evidence generated at one stage, on one instrument, or for one hazard class does not authorize progression in another.

\begin{table*}[h]
\centering
\small
\setlength{\tabcolsep}{3pt}
\renewcommand{\arraystretch}{1.16}
\caption{Evidence-gated stages for extending BLAZE from digital to physical research.}
\label{tab:blaze-deployment-stages}
\newcommand{\blazetablecell}[1]{{\raggedright #1\par}}
\begin{tabular}{@{}p{0.09\textwidth}p{0.20\textwidth}p{0.22\textwidth}p{0.22\textwidth}p{0.19\textwidth}@{}}
\hline
\blazetablecell{\textbf{Stage}} & \blazetablecell{\textbf{Allowed capability}} & \blazetablecell{\textbf{Required evidence}} & \blazetablecell{\textbf{Human authority}} & \blazetablecell{\textbf{Stop condition}} \\
\hline
\blazetablecell{\textbf{D0}\newline Current} & \blazetablecell{Literature review, code, computational experiments, simulation, evidence-linked writing, and review support.} & \blazetablecell{Traceable sources, code, data, configurations, results, and review records.} & \blazetablecell{Approve scope, claims, resource use, authorship, and release.} & \blazetablecell{Failed validation; permission or resource-limit violation.} \\
\hline
\blazetablecell{\textbf{D1}\newline Roadmap} & \blazetablecell{Offline protocol checking without equipment actuation.} & \blazetablecell{D0 audit trail, expert review, hazard analysis, and an offline rejection test.} & \blazetablecell{Approve the protocol and transition toward physical execution.} & \blazetablecell{Unresolved hazard, construct mismatch, or invalid instrument mapping.} \\
\hline
\blazetablecell{\textbf{D2}\newline Roadmap} & \blazetablecell{Supervised actuation through approved tools within predefined limits.} & \blazetablecell{Validated D1 protocol, calibration, interlock tests, anomaly detection, and resource caps.} & \blazetablecell{Approve each run and its limits; control confirmation, stop, and restart.} & \blazetablecell{Anomaly; limit, interlock, or resource-cap breach; missing confirmation.} \\
\hline
\blazetablecell{\textbf{D3}\newline Roadmap} & \blazetablecell{Limited closed-loop adaptation within a pre-authorized low-risk envelope.} & \blazetablecell{Repeated D2 evidence for the specific instrument and hazard class, plus independent safety review.} & \blazetablecell{Approve the adaptation envelope and exceptions; retain stop and restart control.} & \blazetablecell{Envelope breach, drift, safety trigger, or out-of-distribution condition.} \\
\hline
\end{tabular}
\end{table*}

Human authority becomes more important, rather than less, as the system moves along this path. The safety mechanisms listed for D2--D3 overlap with controls discussed for dark laboratories, including interlocks, anomaly detection, bounded closed-loop adaptation, and human stop/restart authority; this overlap is a design relationship, not an implementation claim. Researchers and relevant institutional bodies must classify hazards, approve protocols and stage transitions, define operating limits, authorize exceptions, and control stop and restart decisions. Any future D2--D3 implementation would need to enforce these constraints at the execution layer, outside the agent's own control. Current BLAZE ends at D0; the pre-physical D1 stage and the physical D2--D3 stages are roadmap items rather than reported capabilities. Neither generated protocols nor digital case studies constitute evidence of safe wet-laboratory autonomy or a deployed dark laboratory.

\subsection{Current Deployment Scope}

Because BLAZE spans much of the research lifecycle, its architectural breadth could be mistaken for deployment readiness. A deployment record should therefore specify the research question, named owner, permitted data and tools, resource budget, evidence standard, escalation path, and excluded actions. Within the present boundary, BLAZE supports literature-based research, code development, computational experiments, simulation, evidence-linked drafting, and review support; autonomous wet-laboratory execution, clinical decisions, and other safety-critical physical actions remain outside its scope.

Responsibility remains explicit within this boundary. The principal investigator owns scientific scope and claims, while relevant authorities determine permissible use and preserve approvals as audit artifacts. Final claims, release, authorship, and submission require human approval. Current validation is limited to reported cases and protocols, with broader generalization, governance, and safety-critical deployment requiring further study. The principle is asymmetric: automate reversible, evidence-producing tasks first, but expand authority only through additional validation, tighter constraints, and accountable human ownership. BLAZE distributes scientific work without diffusing responsibility.

\section{Conclusion}
The next transformation of science will not be defined by how much work machines can perform, but by whether scientific inquiry itself can become more cumulative, self critical, and accountable. Reading more papers, running more experiments, and producing manuscripts faster may increase the throughput of research, yet they do not ensure that knowledge advances. Scientific progress depends on a deeper capacity to preserve memory, expose assumptions, confront disagreement, test claims, and revise conclusions in light of failure.
BLAZE addresses this problem by treating the research process, rather than the individual task or agent, as the fundamental unit of scientific intelligence. Its central insight is that intelligence in science is not located in any single model, experiment, or researcher. It emerges from the organized relations among knowledge, hypotheses, evidence, criticism, and judgment. Persistent memory gives continuity to inquiry. Differentiated agents make disagreement explicit and productive. Experiment graphs connect hypotheses to observations and claims, while human governance preserves responsibility for decisions whose scientific and social consequences cannot be delegated. Failed experiments, rejected hypotheses, and unresolved disputes are therefore not residual outputs. They are part of the epistemic structure through which science distinguishes error from discovery and converts experience into knowledge.
BLAZE does not claim that science has already been automated. It advances a more fundamental proposition: scientific autonomy is not a property of an isolated artificial scientist, but of a research organization capable of remembering, questioning, testing, correcting, and governing itself. The future of AI enabled discovery may therefore depend less on building machines that imitate scientists than on building institutions in which human and machine intelligence can form a shared process of inquiry. Such institutions will matter not because they produce results more quickly, but because they preserve the conditions under which results become reliable, contestable, and cumulative knowledge.


\bibliographystyle{unsrt}  
\bibliography{references-ZRP,references-XXX,references-GZP,references-GXY}

\end{document}